%% file: main.tex
\documentclass[10pt,twocolumn,letterpaper]{article}

\usepackage{cvpr}              

\input{preamble}

\definecolor{cvprblue}{rgb}{0.21,0.49,0.74}
\usepackage[pagebackref,breaklinks,colorlinks,allcolors=cvprblue]{hyperref}

\usepackage[accsupp]{axessibility}  

\def\paperID{} 
\def\confName{CVPR}
\def\confYear{2026}

\title{RDFace: A Benchmark Dataset for Rare Disease Facial Image Analysis under Extreme Data Scarcity and Phenotype-Aware Synthetic Generation}

\author{Ganlin Feng$^{1}$ \quad Yuxi Long$^{1}$ \quad Hafsa Ali$^{2}$ \quad Erin Lou$^{3}$ \\ \quad Fahad Butt$^{1}$ \quad Qian Liu$^{4}$ \quad Yang Wang$^{2}$ \quad Pingzhao Hu$^{1,3}$\thanks{Corresponding author (phu49@uwo.ca)} \footnotetext{Code and dataset will be available at \url{https://github.com/Kkathyf/RDFace}.} \vspace{0.3em} \\
{$^1$Western University} \quad
{$^2$Concordia University} \quad
{$^3$University of Toronto} \quad 
{$^4$University of Winnipeg}
}

\begin{document}
\maketitle
\input{sec/0_abstract}    
\input{sec/1_intro}
\input{sec/2_relate}
\input{sec/3_dataset}
\input{sec/4_method}
\input{sec/5_exp_and_result}
\input{sec/6_conclu}

\input{sec/7_ack}

{
    \small
    \bibliographystyle{ieeenat_fullname}
    \bibliography{main}
}


\input{sec/X_supp}

\end{document}

%% file: preamble.tex
\usepackage{longtable}
\usepackage{float}
\usepackage{array}
\usepackage{algorithm}
\usepackage{algorithmic}
\usepackage{tabularx}
\usepackage{multirow}
\usepackage{titletoc}

%% file: sec/0_abstract.tex
\begin{abstract}
Rare diseases often manifest with distinctive facial phenotypes in children, offering valuable diagnostic cues for clinicians and AI-assisted screening systems. However, progress in this field is severely limited by the scarcity of curated, ethically sourced facial data and the high similarity among phenotypes across different conditions. To address these challenges, we introduce \textbf{RDFace}, a curated benchmark dataset comprising 456 pediatric facial images spanning 103 rare genetic conditions (average 4.4 samples per condition). Each ethically verified image is paired with standardized metadata. RDFace enables the development and evaluation of data-efficient AI models for rare disease diagnosis under real-world low-data constraints. We benchmark multiple pretrained vision backbones using cross-validation and explore synthetic augmentation with DreamBooth and FastGAN. Generated images are filtered via facial landmark similarity to maintain phenotype fidelity and merged with real data, improving diagnostic accuracy by up to 13.7\% in ultra-low-data regimes. To assess semantic validity, phenotype descriptions generated by a vision–language model from real and synthetic images achieve a report similarity score of 0.84. RDFace establishes a transparent, benchmark-ready dataset for equitable rare disease AI research and presents a scalable framework for evaluating both diagnostic performance and the integrity of synthetic medical imagery. Project page: \url{https://github.com/Kkathyf/RDFace}.
\end{abstract}

%% file: sec/1_intro.tex
\section{Introduction}
\label{sec:introduction}
Rare diseases (RDs) affect only a limited fraction of individuals in the general population compared to more prevalent diseases. To date, more than 10,000 RDs have been identified, each with a prevalence of 1 in 2000 or less \cite{haendel2020}. Although these disorders are rare individually, nearly 350 million people are affected by RDs worldwide \cite{faviez2020}. The heterogeneity in symptoms, wide range of disorders, limited data, and geographic dispersion all make RD a challenging domain for research \cite{castro2017}. 

Traditional diagnostic pathways for pediatric RDs are often complex and time-consuming, typically involving multiple clinical assessments, genetic testing, and specialist referrals \cite{zurynski2017rare}. As a result, patients frequently experience prolonged diagnostic odysseys before reaching a definitive diagnosis \cite{baynam2016rare}, delaying appropriate treatment and imposing emotional and financial burdens on families and healthcare systems \cite{anderson2013, zurynski2017rare}. These barriers are further amplified in rural or underserved regions, where access to specialized care and timely diagnosis is especially limited \cite{schieppati2008}. A RD Europe study across 17 European countries revealed that 25\% of patients with RDs, such as Duchenne muscular dystrophy and Ehlers-Danlos syndrome, were not properly diagnosed until 5 to 30 years after symptom onset \cite{eurordis2009}. These challenges are especially critical in childhood since many RDs share overlapping symptoms with common childhood conditions, making early clinical recognition difficult \cite{baynam2016rare}.

Notably, many genetic syndromes manifest distinctive craniofacial phenotypes in early childhood, making facial analysis a promising non-invasive signal for RD recognition \cite{qiang2022, kovac2024}. Recent advances in computer vision have demonstrated the potential of facial image analysis for assisting clinical screening and syndrome identification \cite{gurovich2019}. However, progress still remains limited by the lack of standardized benchmarks that reflect realistic clinical scenarios, particularly under extreme data scarcity where only a few samples per condition are available.

\begin{figure*}[ht]
    \centering
    \includegraphics[width=0.85\textwidth]{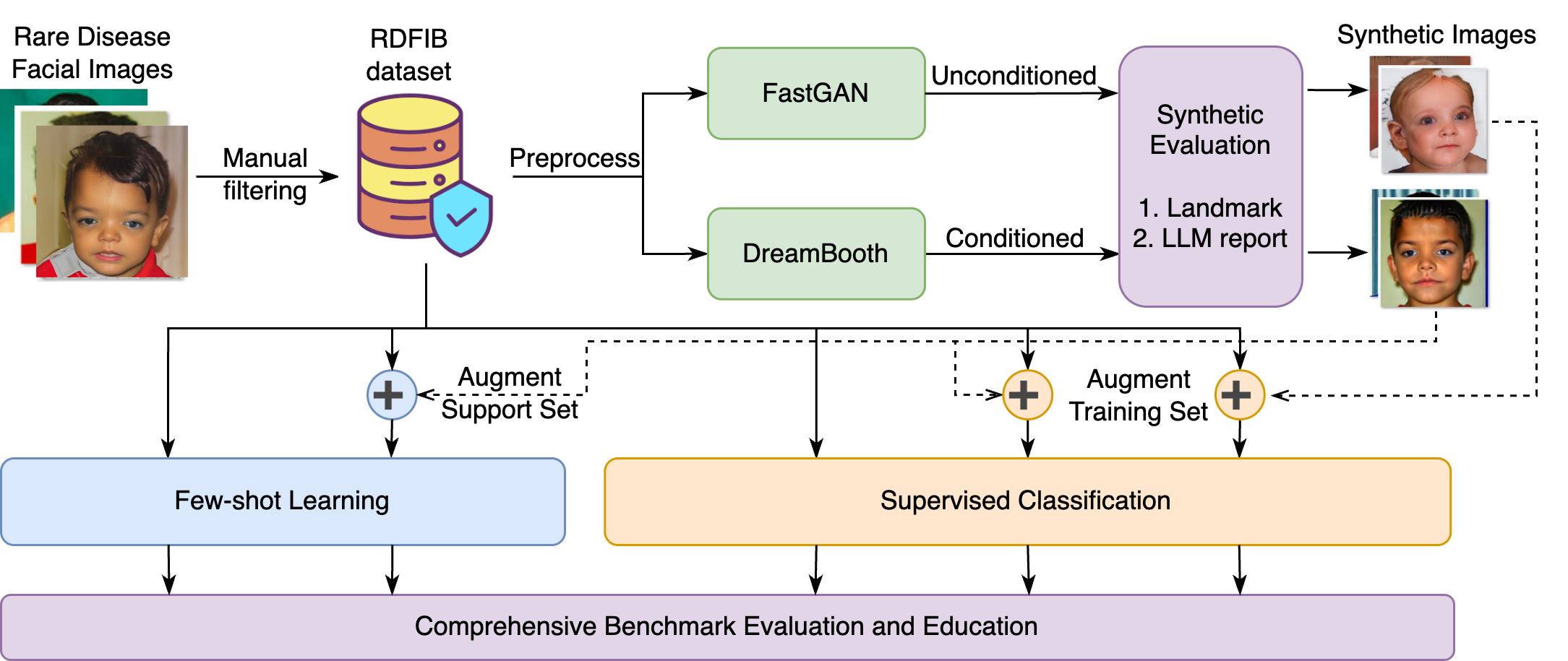}
    \caption{\textbf{Pipeline of RDFace Evaluation.} The pipeline includes dataset curation, classification tasks, synthetic data generation and augmentation, and benchmark evaluation for RD diagnostic support.}
    \label{fig:architecture}
\end{figure*}

To address this gap, we introduce \textbf{RDFace}, a curated benchmark dataset for pediatric RD facial recognition under extreme low-shot conditions. RDFace provides a unified evaluation protocol for systematically comparing learning strategies and analyzing phenotype-aware synthetic augmentation. We consider on three key aspects: (i) benchmark construction under realistic low-data settings, (ii) the effectiveness of phenotype-aligned synthetic augmentation, and (iii) the assessment of generated image fidelity via phenotype-level similarity. The overall evaluation pipeline is illustrated in \Cref{fig:architecture}.

Overall, our contributions are summarized as follows:

\begin{itemize}
\item \textbf{Benchmark Dataset}: We introduce RDFace, a curated pediatric RD facial image benchmark containing 456 images across 103 conditions, where each class contains only 1–7 samples.

\item \textbf{Synthetic Augmentation Study}: Using a unified benchmark protocol, we analyze the effect of phenotype-aligned synthetic facial image augmentation in ultra-low-shot recognition.

\item \textbf{Phenotype-based Fidelity Evaluation}: We propose a phenotype-based evaluation using vision-language models (VLMs) to compare phenotype descriptions from real and synthetic images.
\end{itemize}

%% file: sec/2_relate.tex
\section{Related work}
\label{sec:related_work}
Facial morphology has long supported diagnosis in syndromic disorders \cite{koretzky2016, ren2021}. Advances in deep learning have enabled automated facial phenotyping, bridging genetic medicine and computer vision \cite{germanese2023, yang2021, geremek2021}. Deep learning models have been used to classify genetic syndromes from facial images, but typically focus on under 15 well-represented syndromes with hundreds of images per class \cite{sherif2024, jin2020}. While GestaltMatcher extends to hundreds of syndromes via retrieval-based matching \cite{hsieh2022}, it still requires enough training samples per condition, which is often unrealistic for ultra-RDs.

To cope with data scarcity, few-shot learning has gained traction as an effective strategy \cite{liu2022}. Metric-based methods like Prototypical Networks \cite{snell2017} show promise in medical imaging \cite{Nayem2023}, but remain underexplored for RD facial diagnosis due to high intra-class variability and subtle inter-class differences. Synthetic data generation offers another strategy for augmenting low-resource datasets. While Generative Adversarial Network (GAN) \cite{goodfellow2014} and diffusion models \cite{ho2020} have shown success in other medical contexts \cite{islam2024}, they often require large training sets and fine-tuning. FastGAN improves low-data synthesis via skip-layer excitation and self-supervision \cite{liu2021fastgan}, while DreamBooth is capable of generating high-fidelity samples from 3–5 images \cite{ruiz2023}. In the RD setting, GestaltGAN \cite{kirchhoff2025} enables face synthesis for privacy, but does not evaluate its synthetic images in downstream diagnostic tasks.

Recently, multimodal VLMs have shown promise in linking medical images with clinical semantics for report generation and phenotype interpretation \cite{hartsock2024}. Such models have been applied in radiology and pathology \cite{yildirim2024} to explain predictions and assess image–report consistency, but remain underexplored in RD phenotyping.

%% file: sec/3_dataset.tex
\section{Dataset description}
\label{sec:dataset_description}

\begin{figure*}[!ht]
    \centering
    \includegraphics[width=0.95\textwidth]{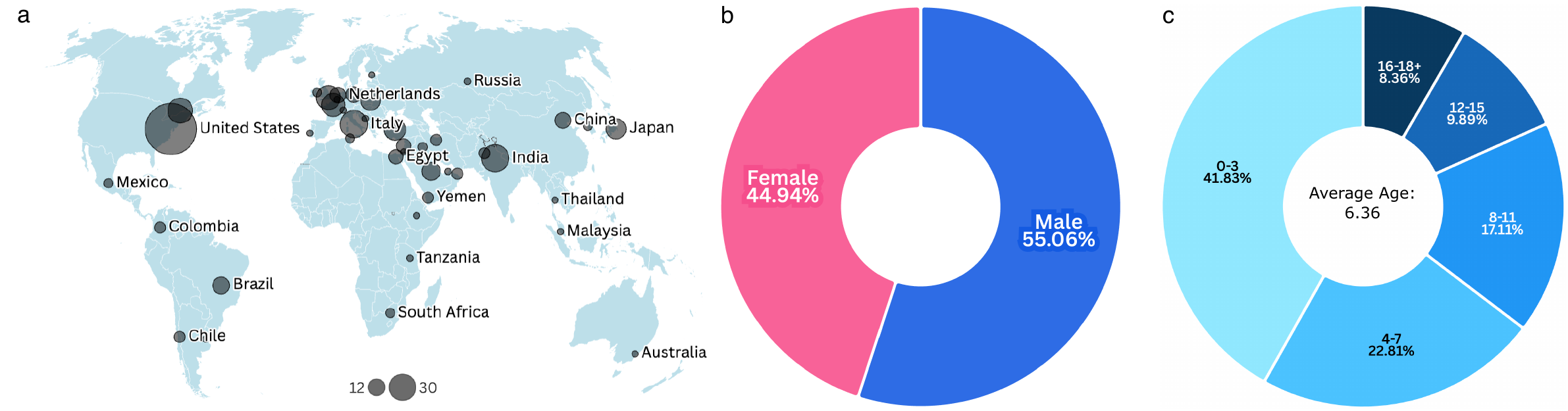}
    \caption{\textbf{Overview of RDFace Dataset.} (a) Geographic distribution of cases across 46 countries, with circle size proportional to the number of cases per country. (b) Sex distribution. (c) Age group distribution of patients with an average age of 6.36 years.}
    \label{fig:RDFace_overview}
\end{figure*}

\subsection{Data collection and sources}
The data collection process was conducted in accordance with institutional research ethics guidelines and received approval from the appropriate Research Ethics Board, ensuring ethical use of publicly sourced clinical imagery. We curated a dataset of 456 frontal pediatric facial portraits spanning 103 rare genetic diseases. Detailed information about each disease is provided in Appendix~\ref{app:metadata}. Candidate conditions were identified from clinical observations in populations with elevated rates of monogenic disorders, often due to historical founder effects and genetic isolation. Although regionally concentrated, many of these diseases are globally recognized and characterized by distinctive craniofacial features, enhancing their relevance for broader research. Each condition was cross-verified with Orphanet \cite{orphanet} to confirm its rarity. For each disease, we recorded metadata including gene association (if known) and a standardized abbreviation for labeling and analysis.

To construct the dataset, we targeted 1-5 high-quality images per disease class, focusing on patients aged 0–18, with emphasis on children under 12 to minimize visual confounding from adult comorbidities \cite{batshaw2014}. Eligible images were required to be frontal-facing, portrait-style, and of sufficient resolution, showing a single patient with open eyes. Images were sourced through structured web searches using disease names and terms such as “face,” “child,” and “patient.” We prioritized peer-reviewed literature, hospital foundations, and verified clinical reports, and included manually reviewed advocacy content when necessary. An overview of the demographic diversity of RDFace is shown in \Cref{fig:RDFace_overview}. More dataset details are in Appendix~\ref{app:dataset_org}.

\noindent\textbf{Expert review.} The dataset curation, creation, and overall design were conducted under the supervision of clinical geneticists to ensure medical plausibility and alignment with diagnostic practice. The real images were initially verified by their original sources, with validation by clinicians explicitly stated in the associated publications. To further ensure dataset reliability, two clinical fellows independently reviewed the plausibility of the image-label associations.

\subsection{Dataset structure}
RDFace is structured as a class-organized image dataset designed for pediatric rare disease classification. The dataset is stored in a directory named \textit{rd\_images/}, which contains 103 subfolders, each corresponding to a rare disease class. Each subfolder is named using the disease abbreviation and contains images representing pediatric patients diagnosed with file names \textit{[disease\_abbr].[index].png}. To facilitate dataset navigation and programmatic access, a metadata file named \textit{disease\_images.csv} is provided in the root directory. This file contains image-level annotations across the following fields: (1) image name, (2) disease name, (3) gene, (4) disease abbreviation, (5) disease subcategory, and (6) Orphanet code. Demographic attributes are summarized at the dataset level but not shared as metadata due to inconsistent availability. The disease abbreviation serves as the class label for all classification tasks conducted in this work.

%% file: sec/4_method.tex
\section{Methodology}
\label{sec:methodology}
This section outlines the benchmarking methodology used in RDFace, encompassing four complementary components: baseline supervised classification, few-shot evaluation, phenotype-aware synthetic data generation, and downstream analysis of synthetic data utility.

\begin{figure*}[t]
    \centering
    \includegraphics[width=0.91\textwidth]{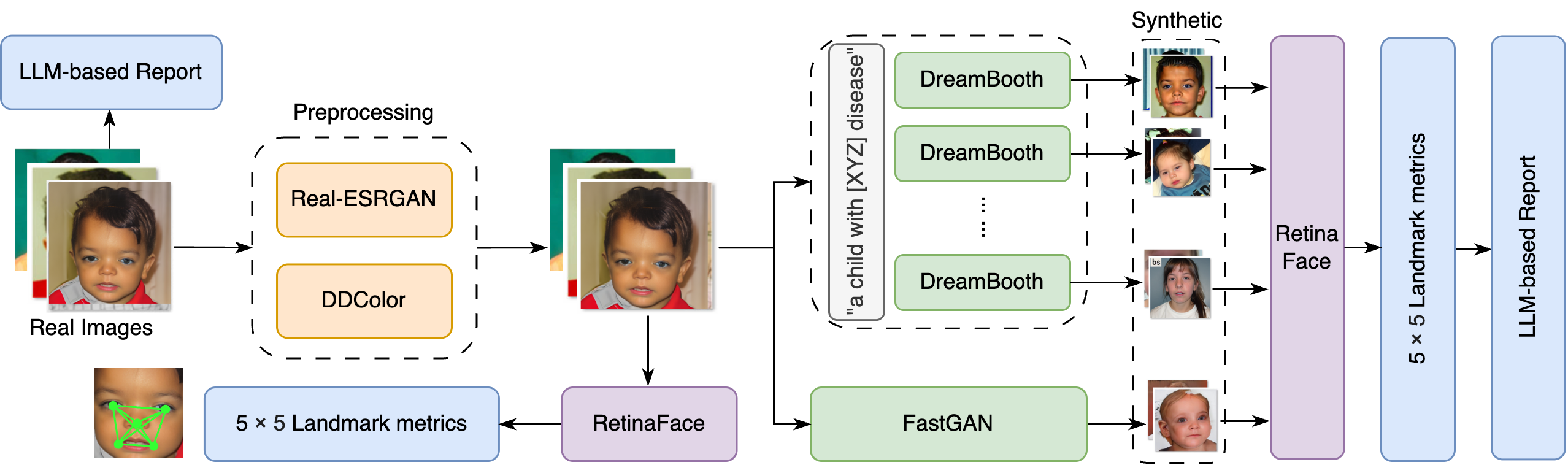}
    \caption{\textbf{Pipeline of synthetic data generation and evaluation.} Real pediatric facial images are first preprocessed using Real-ESRGAN and DDColor, then used to generate synthetic faces via DreamBooth (class-conditioned) and FastGAN (unconditional). Generated images are evaluated for facial realism (RetinaFace and LPIPS) and phenotype consistency (landmark-based cosine similarity).}
    \label{fig:synthetic_generation}
\end{figure*}

\subsection{Standard supervised classification}
\label{sec:methodology_classification}
The RDFace dataset is split into training and testing partitions using a stratified 75\%/25\% image-level split to approximately preserve class distributions. Singleton classes (i.e., those with only one sample) are assigned entirely to the training set to ensure sufficient representation and avoid missing classes during evaluation. During training process, 5-fold cross-validation is performed for hyperparameter tuning, with folds stratified and singleton samples always retained in the training folds.

We benchmark six pretrained backbone architectures: ResNet-152 \cite{he2016}, DenseNet-169 \cite{huang2017}, FaceNet \cite{schroff2015}, VGG-16 \cite{simonyan2015}, Swin Transformer \cite{liu2021}, and CLIP (ViT-B/32) \cite{radford2021}, using weights initialized from ImageNet \cite{deng2009}, VGGFace \cite{parkhi2015}, or VGGFace2 \cite{cao2018}. Input images are resized to $224 \times 224$ and normalized with ImageNet statistics. The final classification layer is replaced with a 103-way softmax corresponding to disease classes. Models are trained using only real training images and evaluated on the held-out test set using Top-$k$ accuracy metrics, where a prediction is considered correct if the true label appears among the model’s top $k$ ranked outputs. Top-$k$ accuracy is the standard evaluation protocol in facial phenotype recognition and rare disease classification benchmarks, ensuring comparability with prior studies.

\noindent\textbf{External architecture comparison.} For external reference, we take DeepGestalt as the representative architecture of prior diagnostic systems. We reproduced its backbone configuration (refered to as the \textit{Gestalt} throughout this paper) using a classifier trained under the RDFace protocol.

\subsection{Few-shot learning}
\label{sec:methodology_fewshot}
We evaluate few-shot learning on RDFace using Prototypical Networks under an $n$-way 1-shot classification setting. Prototypical Network details are provided in Appendix~\ref{app:fewshot_learning}. Singleton classes are excluded, resulting in 99 RD classes. These classes are randomly split into 80\% for training and 20\% for testing, with no class overlap. Within the training set, 5-fold cross-validation is used by holding out a disjoint subset of classes for validation in each fold.

Episodic training is used to match the few-shot evaluation protocol. In each episode, $n$ distinct classes are sampled (e.g., $n=5$), with one support and one query image per class. Prototypes are derived from support embeddings, and queries are classified by their Euclidean distance to these prototypes. The model is optimized using cross-entropy loss computed over query predictions, with gradients back-propagated to update the feature extractor. We train each fold for 600 episodes, using 100 additional episodes for validation.

The same six pretrained backbones used in standard supervised classification are used to encode images. Final performance is evaluated on the meta-test set across three settings: 5-way 1-shot, 10-way 1-shot, and 15-way 1-shot. For each setting, we sample 200 test episodes from unseen classes and report mean accuracy averaged across the five cross-validation folds with standard deviation.

\subsection{Synthetic data generation}
\label{sec:synthetic_generation}
To further address the limited availability of real training data in RD facial classification, we explore synthetic data augmentation through generative modeling. Two approaches are employed: DreamBooth, a supervised diffusion-based model fine-tuned for each disease, and FastGAN, an unsupervised generative adversarial network. These models were selected for their complementary strengths: DreamBooth enables phenotype-aware generation that captures class-specific facial features by conditioning on text and exemplars, while FastGAN offers efficient training on limited data and generates class-agnostic samples to promote overall data diversity. An overview of the generation and evaluation pipeline is illustrated in \Cref{fig:synthetic_generation}. 

To improve image quality, real images are enriched through super-resolution to $512 \times 512$ pixels using Real-ESRGAN \cite{wang2021} and colorization using DDColor \cite{kang2023}. Five key facial landmarks (two eye centers, one nose tip, and two mouth corners) were extracted from preprocessed images using RetinaFace \cite{deng2020}. This five-point configuration is selected for its robustness and consistent detectability across varied image conditions and populations.

\noindent\textbf{DreamBooth.}
Each RD class is fine-tuned separately using its available preprocessed images in the training set, guided by the text prompt "a child with [disease\_abbr] disease". After fine-tuning, 100 synthetic images are generated per disease class. To ensure data quality, outputs flagged as "NSFW" by the model's safety filter are removed.

\noindent\textbf{FastGAN.}
FastGAN is trained unconditionally from scratch on processed training images for 80,000 iterations, saving checkpoints every 10,000 iterations with 1,000 images generated per checkpoint.

\subsection{Synthetic data evaluation protocol} 
\label{sec:synthetic_landmark_similarity} 
Following prior work that jointly evaluates synthetic medical data for both perceptual realism and task-level fidelity \cite{yu2025}, we adopt a dual evaluation framework encompassing (i) intrinsic realism and phenotype fidelity, and (ii) downstream diagnostic utility (\Cref{sec:downstream_analysis}). 
This section focuses on the first component, which evaluates the visual realism and disease relevance of synthetic images.

Facial realism of synthetic images is evaluated via RetinaFace detection scores and LPIPS perceptual similarity \cite{zhang2018} related to real images. RetinaFace assigns a confidence score per image; scores above 0.90 are considered valid, and those exceeding 0.99 are deemed high-quality. Images are symmetrically padded during preprocessing to improve detection for tightly cropped faces.

To assess the disease relevance of synthetic images, we employ a three-part evaluation protocol combining landmark-based similarity, expert clinical review, VLM-assisted phenotype interpretation.

\subsubsection{Landmark-based similarity analysis} 
We first adapt a landmark-based similarity framework to assess phenotypic similarity. For both real and synthetic images, we compute a $5 \times 5$ pairwise Euclidean distance matrix from facial landmarks. An average distance matrix is then derived by aggregating real image matrices for each class. Each synthetic image’s matrix is compared to these class-level averages using cosine similarity. Synthetic images are ranked by similarity to real class prototypes, with lower average rank indicating stronger alignment to true disease phenotypes. After validating this ranking on DreamBooth outputs, we pseudo-label FastGAN-generated images by assigning each to the class with the highest similarity.


\subsubsection{Expert review of synthetic samples}
We conducted an expert review to assess the clinical plausibility of synthetic images to complement automated evaluations. A random subset of images generated by DreamBooth and FastGAN was selected across a range of disease classes. Two medical doctors (MD) were invited to independently evaluate each image using a standardized evaluation form. The primary criterion was visual consistency, whether the facial features in the synthetic image were plausible for the disease label provided. Experts were instructed to consider known clinical characteristics and phenotypic patterns, guided by paired real training images and their own medical knowledge. For each sample, they classified the image as either \textit{Plausible}, \textit{Implausible}, or \textit{Uncertain}.

To quantify inter-rater reliability, we computed evaluation metrics including the percentage of observed agreement, Cohen’s $\kappa$ coefficient, its standard error (SE), and the 95\% confidence interval (CI). These metrics provide an objective measure of annotation consistency and serve as a quality control layer beyond automated scoring and embedding-based similarity analyses.

\subsubsection{VLM-based phenotype description} 
\label{sec:VLMreport_method}
To supplement structural similarity analysis, we further introduce a structured evaluation protocol for assessing the clinical plausibility of synthetic facial images by leveraging the capabilities of VLMs. Specifically, we use the Qwen2.5-VL \cite{bai2025} and LLaVA-NeXT \cite{liu2024llavanext} models with a standardized prompt (Appendix~\ref{app:vlm_prompt}) to generate diagnostic-style clinical reports from both real patient photos and DreamBooth-generated synthetic images.

The facial regions are deliberately chosen to align with the five facial landmarks used in our landmark-based similarity evaluation framework. The model is instructed to describe observations in each region independently and conclude with a predicted diagnosis and a clinical recommendation. This allows us to treat the VLMs as a form of semantic "phenotype reader", capable of interpreting real and synthetic facial morphology in clinical language.

To quantify the similarity between phenotype descriptions, we compute semantic similarity using BioBERT \cite{lee2020} embeddings followed by cosine similarity, yielding region-wise and overall alignment scores for each pair. 

We evaluate similarity across three pair types: real–real, real–synthetic, and synthetic–synthetic. Real–real similarity serves as a reference for the consistency of phenotype descriptions from real data, real–synthetic reflects the fidelity of generated images to real disease characteristics, and synthetic–synthetic provides insight into the consistency of generated samples.

\noindent\textbf{Uncertainty and robustness.}
To assess intra-model stability, we perform a sampling-based uncertainty evaluation using real facial images from RDFace. For each image, the model generates 5 independent reports for each temperature $T \in \{0.7, 0.9, 1.1\}$ using stochastic decoding and then compute pairwise uncertainty $1 - \text{mean similarity}$, where higher values indicate less consistent predictions. We further evaluate cross-model robustness by comparing phenotype descriptions generated by Qwen2.5-VL and LLaVA-NeXT under the same protocol.

\subsection{Downstream analysis of synthetic data utility}
\label{sec:downstream_analysis}
As the second component of the dual evaluation framework, we assess the diagnostic utility of phenotype-aligned synthetic data by injecting them into standard supervised classification and few-shot learning pipelines to quantify performance gains and scaling behavior.

After generating synthetic samples, we evaluate their effectiveness across standard supervised and few-shot learning tasks. Within each disease class, synthetic images are ranked by landmark-based similarity to the real class prototype. We then form augmentation sets of increasing scale (Top-$n$ across all classes; $n\in\{1000,2000,4000,6000\}$) to study scaling behavior. To verify that landmark-selected samples also maintain visual realism, we compute the average RetinaFace confidence score and LPIPS perceptual similarity across each Top-$n$ subset, ensuring the augmented data are both phenotype-consistent and visually plausible.

\noindent\textbf{Supervised scaling effect.}
To analyze the scaling effect of synthetic augmentation in standard supervised classification, models are retrained under a consistent protocol using (i) real data only (as baselines) and (ii) real data combined with synthetic subsets of increasing size (1000-6000). Synthetic images are excluded from the test set to preserve evaluation integrity. This setup enables a controlled analysis of how scaling phenotype-aligned synthetic data impacts classification performance.

\noindent\textbf{Few-Shot learning.}
In few-shot learning, synthetic samples are added to the support sets to increase training diversity without altering the few-shot structure. For each class with $m$ real support examples, $(10-m)$ synthetic images are sampled to reach 10 support examples per class. This augmentation strategy enables the exploration of different support set sizes during training. Testing is performed strictly under the original $n$-way 1-shot configuration, using only real images for support and query sets to ensure unbiased evaluation.

%% file: sec/5_exp_and_result.tex
\section{Experiments and results}
This section presents the experimental evaluation of supervised classification, few-shot learning, and synthetic data augmentation on the RDFace dataset. Detailed hyperparameter settings and error bar calculations are provided in Appendix~\ref{app:training} and Appendix~\ref{app:stats_reporting} respectively.

\subsection{Standard supervised classification}
We first evaluated supervised classification across six backbone architectures using real facial images only. As shown in \Cref{tab:supervised_classification_results}, performance varied substantially across models. DenseNet achieved the highest Top-1 accuracy at 15.93\%, followed by Swin Transformer (14.34\%) and VGG (11.68\%). Other models, such as ResNet, FaceNet, and CLIP, trailed behind with lower accuracy, particularly CLIP (3.01\%), which underperformed across all Top-$k$ metrics.

\setlength{\tabcolsep}{2pt}
\begin{table}[h]
\caption{Standard supervised classification results using real training data across different backbones and baseline Gestalt.\protect\footnotemark}
\small
\centering
\begin{tabular}{l|cccc}
\toprule
\textbf{Acc (\%)} & \textbf{Top-1} & \textbf{Top-5} & \textbf{Top-10} & \textbf{Top-30} \\
\midrule
ResNet   & 6.90 (1.45)  & 18.58 (3.00) & 28.50 (3.67) & 54.34 (2.39) \\
DenseNet & \textbf{15.93} (2.34) & \textbf{33.63} (3.70) & \textbf{43.01} (2.63) & \textbf{64.42} (1.92) \\
FaceNet  & 9.91 (1.81)  & 24.60 (5.43) & 34.87 (4.75) & 58.23 (5.06) \\
VGG      & 11.68 (1.58) & 29.91 (2.68) & 38.41 (1.34) & 60.88 (2.02) \\
Swin-T   & 14.34 (2.61) & 26.19 (2.04) & 35.93 (3.17) & 58.41 (3.81) \\
CLIP     & 3.01 (1.48)  & 12.74 (2.84) & 19.12 (5.18) & 42.30 (4.40) \\
Gestalt & 6.19 (1.40) & 17.52 (1.58) & 27.79 (3.68) & 49.20 (2.91)
\\
\bottomrule
\end{tabular}
\label{tab:supervised_classification_results}
\end{table}

\footnotetext{All results in this paper are reported as mean (standard deviation) unless otherwise noted. 
The best results are highlighted in \textbf{bold}.}

Top-5 and Top-10 accuracy results further reveal that model performance improves significantly when more guesses are permitted. For example, DenseNet’s Top-5 accuracy reached 33.63\%, and Top-30 accuracy exceeded 64\%. This suggests that while Top-1 classification remains difficult in ultra-low-shot settings, models can still learn useful representations for phenotype-based narrowing.

\subsection{Few-shot learning}
Prototypical Networks were applied under 5, 10, and 15-way 1-shot settings (see \Cref{tab:fewshot_real_only_results}). Few-shot learning achieved substantially higher accuracy within its simplified episodic setup, highlighting its adaptability in ultra-low-shot settings. DenseNet achieved the highest accuracy in the 5-way configuration (26.20\%), while ResNet maintained more stable performance as task complexity increased (18.15\% and 17.99\% for 10- and 15-way, respectively). CLIP and FaceNet were less effective, especially in higher-way tasks, reflecting limited generalization of their embeddings in RD phenotyping.

\setlength{\tabcolsep}{5pt}
\begin{table}[h]
\caption{Few-shot learning results under different settings using real training data across different backbone models.}
\small
\centering
\begin{tabular}{l|ccc}
\toprule
\textbf{Acc (\%)} & \textbf{5-way 1-shot} & \textbf{10-way 1-shot} & \textbf{15-way 1-shot} \\
\midrule
ResNet   & 24.18 (2.56) & \textbf{18.15} (2.58) & \textbf{17.99} (1.26) \\
DenseNet & \textbf{26.20} (2.01) & 17.36 (1.66) & 17.30 (3.45) \\
FaceNet  & 25.16 (4.89) & 12.93 (2.23) & 8.03 (1.40) \\
VGG      & 21.54 (3.72) & 10.82 (2.14) & 9.05 (2.20) \\
Swin-T   & 22.24 (2.88) & 10.94 (2.59) & 8.03 (2.41) \\
CLIP     & 23.48 (5.28) & 12.33 (1.96) & 7.30 (1.90) \\
\bottomrule
\end{tabular}
\label{tab:fewshot_real_only_results}
\end{table}

\subsection{Synthetic data generation and evaluation}

Representative synthetic images for Aarskog-Scott syndrome (AAR) generated by DreamBooth, together with unconditional samples from FastGAN, are presented in \Cref{fig:synthetic_comparison_aar}. Additional examples generated by both methods are provided in Appendix~\ref{app:synthetic_images}. Overall, the generative process resulted in 10,300 DreamBooth-generated and 8,928 FastGAN-generated synthetic images.

\begin{figure}[h]
    \centering
    \begin{subfigure}[b]{0.45\columnwidth}
        \centering
        \includegraphics[height=0.9\columnwidth]{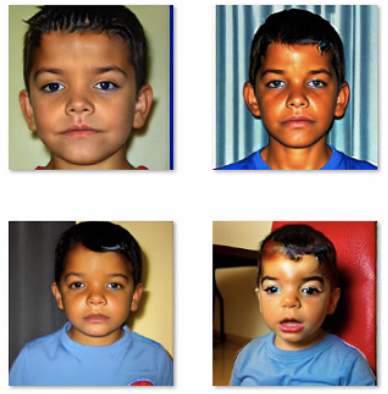}
        \caption{DreamBooth}
        \label{fig:aar_dreambooth}
    \end{subfigure}
    \hspace{0.05\columnwidth}
    \begin{subfigure}[b]{0.45\columnwidth}
        \centering
        \includegraphics[height=0.9\columnwidth]{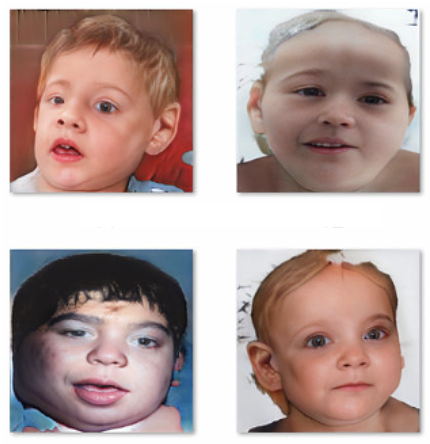}
        \caption{FastGAN}
        \label{fig:aar_fastgan}
    \end{subfigure}
    \caption{\textbf{Representative samples of synthetic images.} (a) DreamBooth-generated synthetic images conditioned on AAR. (b) FastGAN-generated synthetic images.}
    \label{fig:synthetic_comparison_aar}
\end{figure}

\begin{figure*}[t]
    \centering
    \includegraphics[width=0.94\textwidth]{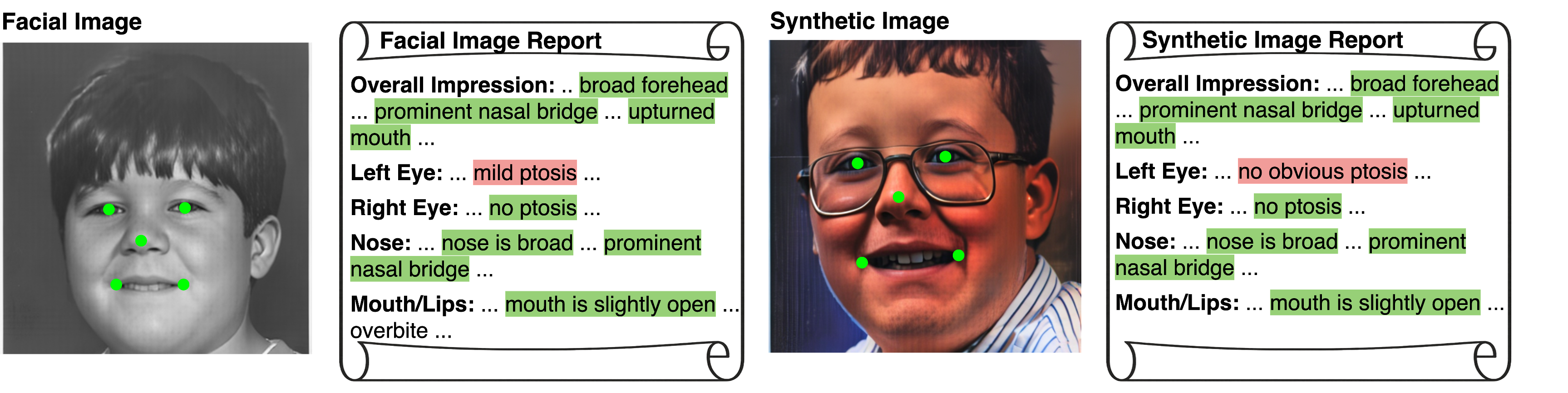}
    \caption{\textbf{Comparison of phenotype descriptions generated by VLM between a real image and one corresponding synthetic image.} Left: real image; Right: DreamBooth-generated synthetic image. The real image has been visually processed to reduce identifiability in accordance with privacy and ethical considerations. Green indicates consistent phenotype terms while red indicates conflicting descriptions.}
    \label{fig:bar_example}
\end{figure*}

Beyond visual inspection, we evaluate the visual realism of generated samples. After filtering out samples flagged by safety checks or with RetinaFace confidence below 0.90, 99.42\% of DreamBooth and 99.70\% of FastGAN samples achieve high detection confidence ($>$ 0.99). Perceptual similarity to real training images, measured using LPIPS, was 0.5337 for FastGAN and 0.4871 for DreamBooth.

Phenotypic alignment was assessed via landmark-based cosine similarity and VLM-generated phenotype reports. DreamBooth-generated images achieved an average landmark similarity rank of 19.74, confirming structural consistency with real disease prototypes. Additional details on the landmark similarity analysis can be found in Appendix~\ref{app:landmark-similarity}.

\noindent\textbf{Expert review of synthetic samples.}
To validate the clinical relevance of the synthetic images, a random subset of 50 DreamBooth and 50 FastGAN samples was evaluated. DreamBooth images were marked as \textit{Plausible} in 62–76\% of cases, depending on whether at least one or both reviewers agreed, while FastGAN images reached only 2–38\%. Inter-rater reliability further supported these findings. For DreamBooth, the observed agreement was 84.0\% with Cohen’s $\kappa = 0.65$, indicating substantial agreement; while FastGAN showed only 38.0\% observed agreement with $\kappa = 0.07$, suggesting poor consistency. Detailed ratings are summarized in Appendix~\ref{app:expert_review}. The higher plausibility of DreamBooth outputs suggests that conditioning helps preserve phenotype-specific features for clinical interpretation.

\begin{figure}[h]
    \centering
    \includegraphics[width=0.76\columnwidth]{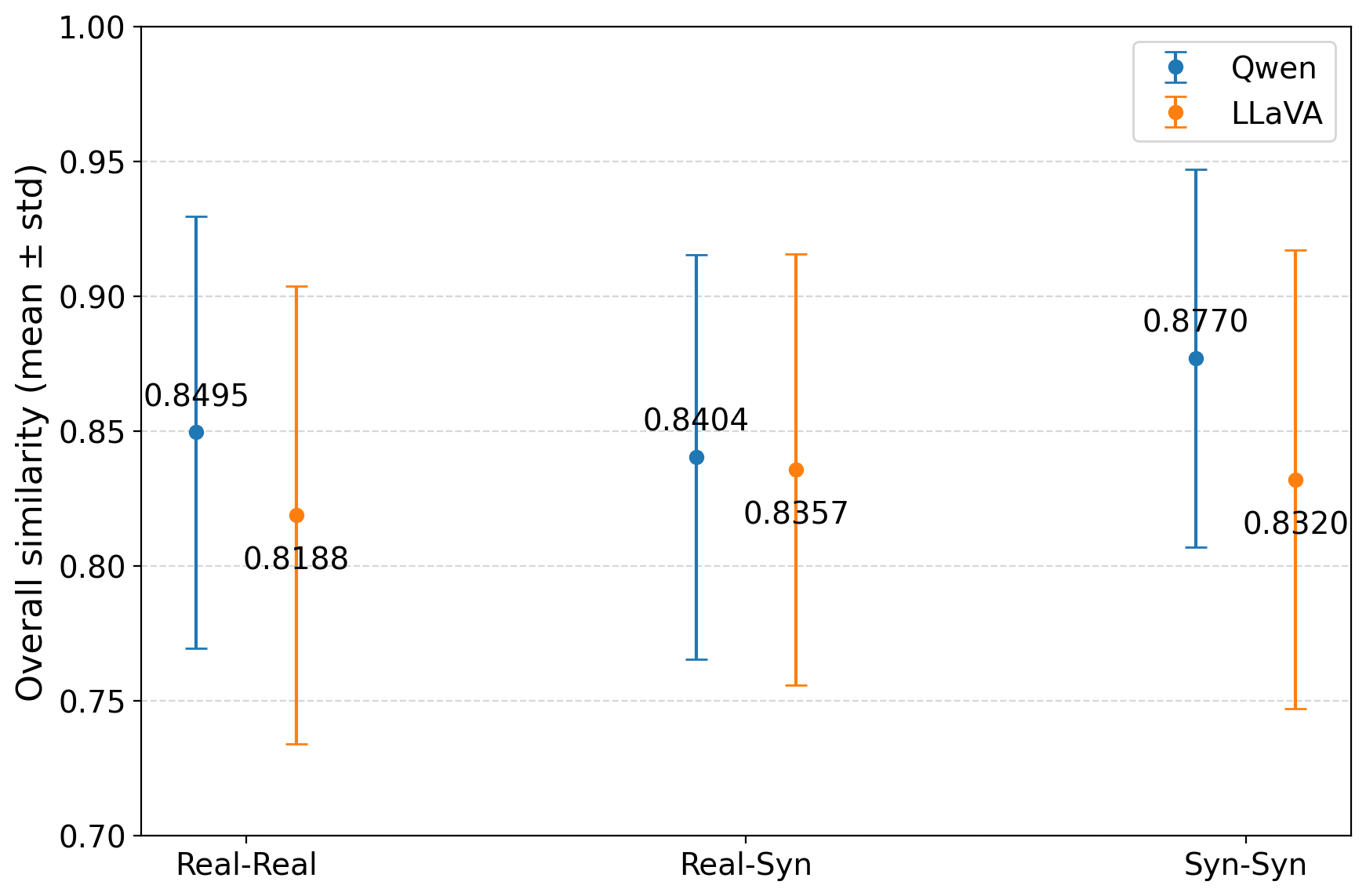}
    \caption{\textbf{Overall similarity scores for Qwen and LLaVA across different comparisons.} Error bars indicate standard deviation.}
    \label{fig:2vlms}
\end{figure}

\noindent\textbf{VLM-cased phenotype alignment.}
The example in \Cref{fig:bar_example} shows consistent semantic alignment between real and synthetic descriptions, with most discrepancies occurring in less distinctive regions. More cases are provided in Appendix~\ref{app:vlm_eval}. We further evaluate phenotype alignment using VLM-generated reports across different image pairs. As shown in \Cref{fig:2vlms}, real-syn similarity is comparable to real-real across both models, indicating that synthetic images preserve disease-specific phenotype information. Syn-syn similarity is consistently high, suggesting stable phenotype representations among generated samples. Together, these results support the utility of DreamBooth samples.

\noindent\textbf{Uncertainty and robustness.} Via stochastic sampling, the stability of Qwen-generated reports yields a mean uncertainty score of $0.108 \pm 0.017$, indicating high consistency across samples. As shown in Appendix~\ref{app:vlm_cross_model}, cross-model similarity remains consistent across real and synthetic images, indicating that phenotype descriptions are not strongly dependent on the choice of VLM.

\noindent\textbf{Potential regional bias.} We further analyze potential bias across geographic regions on standard supervised learning and phenotype report similarity. Results indicate consistent trends across regions and are provided in the Appendix~\ref{app:regional_bias}.

\subsection{Impact of synthetic data augmentation}
We evaluated the impact of synthetic data generated by DreamBooth and FastGAN across standard supervised and few-shot classification tasks under multiple configurations.

For standard supervised classification, we first compare real only performances with generic augmentation methods as baselines. On our best backbone DenseNet: MixUp \cite{zhang2018mixup} and CutMix \cite{cutmix2019} achieve 15.75\% and 16.11\% Top-1 accuracy respectively, suggesting generic augmentations provide limited or inconsistent gains under extreme low-shot conditions.
DreamBooth (DB) augmentation (\Cref{tab:synthetic_classification_results}) consistently improved Top-$1$ accuracy across nearly all backbones, 
demonstrating the benefit of phenotype-aligned synthesis. For example, DenseNet increased from 
15.93\% (real only) to 17.52\% with DB augmentation, and VGG from 11.68\%  to 16.64\%. In contrast, FastGAN (FG) augmentation 
led to performance degradation across most models. For instance, DenseNet dropped from 15.93\% 
to 13.27\%. The mixed (Real + DB + FG) configuration partially recovered performance, reaching intermediate accuracy between the two augmentations and being slightly better than the real-only baselines. These results show that class-conditioned augmentation is consistently beneficial, while unconditioned synthesis can distort class distributions.

\setlength{\tabcolsep}{1.5pt}
\begin{table}[h]
\caption{Standard supervised classification results (Top-1 accuracies) under landmark-based Top-1000 synthetic data augmentation across different backbone models.}
\centering
\small
\begin{tabular}{l|cccc}
\toprule
\textbf{ACC (\%)} & \textbf{Real only} & \textbf{Real+DB} & \textbf{Real+FG} & \textbf{Real+DB+FG} \\
\midrule
ResNet   & 6.90 (1.45)  & 12.21 (1.70) & 8.50 (1.48) & 8.32(1.73)\\
DenseNet & \textbf{15.93} (2.34) & \textbf{17.52} (2.29) & \textbf{13.27} (3.13) & \textbf{16.46} (2.63)\\
FaceNet  & 9.91 (1.81)  & 15.04 (2.87) & 6.55 (2.48) & 10.97 (2.55)\\
VGG      & 11.68 (1.58) & 16.64 (4.07) & 7.26 (2.37) & 12.92(2.84) \\
Swin-T   & 14.34 (2.61) & 16.81 (1.40) & 10.44 (1.70) & 14.34 (1.70)\\
CLIP     & 3.01 (1.48)  & 9.03 (2.29)  & 1.42 (1.34) & 4.25 (1.58)\\
Gestalt  & 6.19 (1.40)  & 9.03 (1.45)  & 3.19 (1.84) & 5.31 (0.88) \\
\bottomrule
\end{tabular}
\label{tab:synthetic_classification_results}
\end{table}


\noindent\textbf{Ablation study on scaling effect.} To analyze how the quantity and type of synthetic data influence model performance, we varied the number of DreamBooth- and FastGAN-generated samples added per class. As shown in Appendix~\ref{app:downstream_results_classification}, both models exhibit distinct scaling behaviors. DreamBooth shows a clear non-linear improvement pattern: accuracy rises sharply between the Top-1000 and Top-4000 subsets and then plateaus or slightly declines at Top-6000. For instance, DenseNet’s Top-1 accuracy increases from 15.93\% (real only) to 17.52\% (Top-1000) and reaches 21.06\% at Top-6000, indicating that moderate quantities of phenotype-aligned samples most effectively enhance generalization. In contrast, FastGAN displays an immediate and persistent decline as more samples are added, reflecting low signal-to-noise in its unconditioned generation. Overall, the ablation reveals that performance gains saturate with excessive synthetic data and are driven by phenotype fidelity rather than dataset volume, underscoring the value of phenotype-aware generation.

Few-shot learning results (see \Cref{tab:synthetic_fewshot_results} and Appendix~\ref{app:downstream_results_few_shot}) further validate these observations. DreamBooth augmentation outperformed real-only baselines in most 1-shot settings. For instance, DenseNet improved from 26.20\% to 29.88\% (5-way 1-shot), and ResNet improved from 24.18\% to 25.72\%. Gains were also observed in 5-shot settings, where larger support sets enabled better utilization of synthetic diversity (e.g., ResNet reaching 33.62\%). CLIP, however, showed modest or inconsistent benefits, possibly due to its weaker initial alignment with facial phenotypes.

\setlength{\tabcolsep}{10pt}
\begin{table}[h]
\caption{Few-shot learning results under synthetic data augmentation across different backbone models.}
\centering
\small
\begin{tabular}{l|cc}
\toprule
\textbf{5-way 1-shot ACC (\%)} & \textbf{Real only} & \textbf{Real + DB} \\
\midrule
ResNet   & 24.18 (2.56) & 25.72 (1.62) \\
DenseNet & \textbf{26.20} (2.01) & \textbf{29.88} (1.51) \\
FaceNet  & 25.16 (4.89) & 23.60 (4.06) \\
VGG      & 21.54 (3.72) & 21.02 (5.85) \\
Swin-T   & 22.24 (2.88) & 26.72 (4.34) \\
CLIP     & 23.48 (5.28) & 22.30 (2.63) \\
\bottomrule
\end{tabular}
\label{tab:synthetic_fewshot_results}
\end{table}

Supplemental visual quality analysis (Appendix~\ref{app:synthetic-metrics}) confirmed that Top-$n$ landmark similarity ranking is predictive of image realism, which explains why DreamBooth samples generalize better across models and settings.

%% file: sec/6_conclu.tex
\section{Conclusion}
\label{sec:discussion}
\noindent\textbf{Findings and implications.}
Overall, these findings highlight the versatility of the proposed rare disease dataset in supporting diverse analytical tasks, from classification and few-shot learning to phenotype evaluation. Supervised and few-shot classification results underscore the difficulty of ultra-low-shot rare disease diagnosis, with few-shot methods offering modest gains in constrained settings. Synthetic data, particularly DreamBooth-generated samples, effectively mitigated data scarcity. Their consistent performance gains suggest improvements stem from higher phenotype fidelity rather than sample quantity or overfitting. The landmark-based similarity offers interpretable grounded validation and enable pseudo-labelling. Finally, the VLM-generated diagnostic reports demonstrated strong semantic consistency and interpretability, underscoring their potential for future clinical and educational applications.

\noindent\textbf{Limitations.}
RDFace reflects the real-world scarcity and imbalance inherent in rare disease data and, while currently limited in scale, provides a valuable foundation for studying generalization under such constraints. The images were collected from heterogeneous online sources with varying completeness of demographic metadata, which may limit the extent of bias or cross-population generalizability.

\noindent\textbf{Conclusion.}
RDFace is a curated benchmark dataset designed for rare disease facial analysis under real-world scenarios. It provides standardized settings for evaluating supervised, few-shot, and phenotype-aware learning methods, enabling systematic assessment of model performance under data scarcity. By integrating real pediatric images with synthetic augmentations and enabling both structural and semantic evaluations, RDFace supports comprehensive and reproducible assessment of model behavior. Our experiments demonstrate that phenotype-aligned synthetic data improves recognition while preserving clinically relevant features. Together, RDFace thus dserves as a foundation for developing reliable AI systems for rare disease diagnosis.

%% file: sec/7_ack.tex
\section*{Acknowledgement}
This work was supported in part by the Canada Research Chairs Tier II Program (CRC-2021-00482) and the Canada Foundation for Innovation John R. Evans Leaders Fund (JELF) program (\#43481). All data collection procedures were approved by the Western University Health Science Research Ethics Board (HSREB) (Reference No. 2023-122744-77394). Facial photographs of children with rare diseases were collected from publicly available sources, including the published literature and foundation websites, and the authors gratefully acknowledge these sources. The authors sincerely thank Dr. Patrick Frosk for his support in the design of the project. The authors also acknowledge the Digital Research Alliance of Canada and Compute Canada for providing the computational resources used in this study.

%% file: sec/X_supp.tex
\clearpage
\appendix
\setcounter{page}{1}
\maketitlesupplementary

\setcounter{section}{0}
\renewcommand{\thesection}{\Alph{section}}

\renewcommand{\thetable}{S\arabic{table}}
\renewcommand{\thefigure}{S\arabic{figure}}
\setcounter{table}{0}
\setcounter{figure}{0}

\begin{center}
    {\Large {Appendix Contents}}
\end{center}
\vspace{0.5em}

\startcontents[appendix]
\printcontents[appendix]{}{1}{}

\clearpage

\section{Dataset documentation}
\label{app:dataset}
\subsection{Disease list and metadata}
\label{app:metadata}
We provide a complete list of the 103 rare disease classes included in the RDFace dataset. As summarized in \Cref{tab:s1_metadata}, each entry includes the disease name, abbreviation (Abbr) (used for labeling), associated gene (if available), clinical subcategory based on Orphanet (if available), Orphanet code, and the number of real facial images (\# Img) curated for that class.

\footnotesize 
\setlength{\tabcolsep}{10pt} 
\renewcommand{\arraystretch}{1.2} 

\begin{longtable}{
    p{4cm}    
    >{\centering\arraybackslash}p{0.8cm}  
    >{\centering\arraybackslash}p{2.2cm}  
    p{3.4cm}    
    >{\centering\arraybackslash}p{2.1cm}  
    >{\centering\arraybackslash}p{1.0cm}  
}
\caption[Metadata of rare disease classes in the RDFace dataset.]%
{Metadata of rare disease classes in the RDFace dataset.\footnotemark}
\label{tab:s1_metadata} \\
\toprule
\textbf{Disease Name} & \textbf{Abbr} & \textbf{Gene} & \textbf{Subcategory} & \textbf{Orphanet Code} & \textbf{\# Img} \\
\midrule
\endfirsthead

\toprule
\textbf{Disease Name} & \textbf{Abbr} & \textbf{Gene} & \textbf{Subcategory} & \textbf{Orphanet Code} & \textbf{\# Img} \\
\midrule
\endhead

\midrule
\endfoot

\bottomrule
\endlastfoot
Aarskog-Scott syndrome  & AAR & FGD1 & Delayed puberty  & 915  & 5  \\
Aicardi-Goutieres & AIC & TREX1 & ? & 51  & 5  \\
Triple A syndrome & ALL & AAAS & multisystem disease & 869  & 5  \\
Allan-herndon-dudley Syndrome & ALLA & SLC16A2 & Ataxia & 59  & 5  \\
Alport syndrome & ALP & ? & Abnormal retinal morphology & 63  & 4  \\
Alpha-thalassemia & ALPH & ? & Cholestasis & 846  & 5  \\
Alpha-mannosidase deficiency & ALPM & MAN2B1 & lysosomal storage disease & 61  & 5  \\
Alstrom & ALS & ALSM1 & multisystemic disorder & 64  & 5  \\
Angelman Syndrome  & ANG & UBE3A & Intellectual disability & 72  & 5  \\
X-linked cleft palate and ankyloglossia & ANK & TBX22 & developmental defect during embryogenesis syndrome & 324601  & 2  \\
Apert syndrome & APE & FGFR2  & Cardiomyopathy & 87  & 5  \\
AR Polycystic Kidney Disease & ARP & PKHD1 & hepatorenal fibrocystic syndrome  & 731  & 1  \\
Arterial Tortuosity & ART & ? & connective tissue disorder & 3342  & 5  \\
Ataxia Telangiectasia & ATA & ATM & combined dystonia & 100  & 2  \\
Atypical Rett Syndrome & ATYP & MECP2, GABBR2, STXBP1, CDKL5 & Calcium nephrolithiasis & 3095  & 5  \\
Auriculocondylar Syndrome & AURI & EDN1, PLCB4, GNAI3 & Abnormal soft palate morphology to abnormality of the uvulva & 137888  & 5  \\
Bainbridge-ropers Syndrome  & BAI & ASXL3 & Severe postnatal growth retardation & 352577  & 5  \\
Bardet-Biedl & BAR & BBS2 & ciliopathy with multisystem involvement & 110  & 5  \\
Barber-say Syndrome & BARB & TWIST2 & Hyperextensible skin & 1231  & 5  \\
Barth syndrome & BART & ? & Cardiomyopathy & 111  & 5  \\
Beckwith-wiedemann Syndrome & BEC & ? & Multiple renal cysts & 116  & 5  \\
Bohring-opitz Syndrome & BOH & ASXL1 & Muscular hypotonia & 97297  & 5  \\
Boycott-Beaulieu-Innes & BOY & THOC6 & syndromic intellectual disability disorder & 363444  & 5  \\
Congenital adrenal hyperplasia  & CAH & ? & Congenital adrenal hyperplasia & 418  & 5  \\
Canavan Disease & CAN & ASPA & Abnormality of serum amino acid level & 141  & 5  \\
COFS syndrome & CER & ERCC6 & diseases of DNA repair & 1466  & 5  \\
Chudley-McCullough & CHU & GPSM2 & syndromic deafness & 314597  & 4  \\
Cleidocranial Dysplasia & CLE & CBFA1 & developmental abnormality of bone & 1452  & 5  \\
Clouston Syndrome & CLOU & GJB6 & Ectodermal dysplasia & 189  & 4  \\
CODAS & COD & LONP1 & multiple congenital anomalies syndrome  & 1458  & 5  \\
Coffin-lowry Syndrome & COF & RPS6KA3 & Spasticity & 192  & 5  \\
Combined pituitary hormone deficiencies, genetic forms & COM & PROP1 & Congenital hypopituitarism & 95494  & 4  \\
SLC39A8-CDG & COND & SLC39A8 & ? & 468699  & 5  \\
Cranioectodermal dysplasia & CRA & DPH1 & developmental disorder & 1515  & 5  \\
Crisponi Syndrome & CRIS & CLCF1, CRLF1 & Malignant hyperthermia & 1545  & 5  \\
3C syndrome & CSY & CCDC22, WASHC5 & ? & 7  & 5  \\
Cushing disease & CUS & USP8, CDH23 & Abnormal bleeding & 96253  & 4  \\
Cystic Fibrosis & CYS & CFTR & ? & 586  & 5  \\
Diamond-blackfan Anemia  & DIA & RPS19 & Colon cancer & 124  & 5  \\
Donnai-barrow Syndrome & DON & LRP2 & Partial agenesis of the corpus callosum & 2143  & 5  \\
DOORS syndrome & DOO & TBC1D24 & Hypothyroidism & 79500  & 5  \\
Dopa-Responsive Dystonia & DOP & TH & group of diseases & 255  & 3  \\
Dysosteosclerosis & DYS & ? & primary bone dysplasia disease & 1782  & 2  \\
Early infantile epileptic encephalopathy & EPIE & ? & epileptic encephalopathy & 1934  & 5  \\
Fanconi Anemia & FANC & FANCC & DNA repair disorder  & 84  & 5  \\
Geroderma Osteodysplastica & GEO & GORAB & ? & 2078  & 5  \\
Glutaryl-CoA dehydrogenase deficiency & GLU & GCDH & neurometabolic disorder & 25  & 5  \\
HHH (hyperornithinemia-hyperammonemia-homocitrullinuria) & HHH & SLC25A15 & disorder of urea cycle metabolism  & 415  & 1  \\
Hypohidrotic Ectodermal Dysplasia & HYPE & EDA1 & disorder of ectoderm development & 238468  & 1  \\
Hypophosphatasia & HYPO & ALPL & metabolic disorder  & 436  & 5  \\
Severe combined immunodeficiency & IMMU & DCLRE1C & primary immunodeficiency & 183660  & 5  \\
Joubert Syndrome & JOUA & TMEM237 & ? & 475  & 5  \\
Juvenile Amyotrophic Lateral Sclerosis & JUV & ALS2, HNRNPA2B1, HNRNPA1 & Motor neuron atrophy & 300605  & 5  \\
Kawasaki Disease & KAW & ? & Arrhythmia & 2331  & 5  \\
Infantile Krabbe Disease & KRA & GALC, PSAP & Generalized myoclonic seizure & 206436  & 5  \\
Laron Syndrome & LAR & GHR & Hypoglycemia & 633  & 3  \\
Leigh & LEI & NDUFV1 & progressive neurological disease  & 506  & 5  \\
Leprechaunism & LEP & INSR & Hypoglycemia & 508  & 7  \\
Loeys-Dietz & LOE & TGFB2 & connective tissue disorder  & 60030  & 5  \\
Malignant hyperthermia of anesthesia & MAL & RYR1 & pharmacogenetic disorder of skeletal muscle & 423  & 3  \\
Maple Syrup Urine Disease & MAP & BCKDHB & disorder of branched-chain amino acid metabolism & 511  & 5  \\
Marden-walker Syndrome & MAR & PIEZO2 & Muscular dystrophy & 2461  & 5  \\
Marinesco-sjögren Syndrome & MARI & INPP5K, SIL1 & Muscular dystrophy & 559  & 5  \\
Oculotrichoanal syndrome & MBO & FREM1 & multiple congenital anomalies & 2717  & 5  \\
Mucopolysaccharidosis type 4 & MOR & ? &  lysosomal storage disease & 582  & 5  \\
Mowat-wilson Syndrome & MOW & ZEB2 & Abdominal distention  & 2152  & 5  \\
Mucopolysaccharidosis type 7 & MPS & GUSB & lysosomal storage disease & 584  & 5  \\
Multiple Sulfatase Deficiency & MUL & SUMF1 & Progressive neurologic deterioration & 585  & 5  \\
Ochoa Syndrome & OCH & HPSE2, LRIG2 & Renal insufficiency & 2704  & 5  \\
Oculocutaneous Albinism & OCU & TYR & disorders of melanin biosynthesis & 55  & 5  \\
Odonto-onycho-dermal dysplasia & ODO & WNT10A & ectodermal dysplasia & 2721  & 5  \\
Osteogenesis imperfecta & OST & SEPINF1 & group of diseases & 666  & 5  \\
Parietal Foramina & PAR & ? & ? & 60015  & 3  \\
Glycogen storage disease due to acid maltase deficiency, late-onset & POM & GAA & Glycogen storage disease & 420429  & 5  \\
Pontocerebellar Hypoplasia & PON & TOE1 & ? & 98523  & 5  \\
Primary Hyperoxaluria & PRI & GRHPR & disorder of glyoxylate metabolism & 416  & 2  \\
Microcephaly-lymphedema-chorioretinopathy syndrome & PRIM & ? & ? & 2526  & 5  \\
Propionic Acidemia & PRO & PCCB & organic aciduria  & 35  & 5  \\
PRUNE1-related neurological syndrome & PRU & PRUNE1 & ? & 544469  & 2  \\
Pten Hamartoma Tumor Syndrome & PTEN & ? & Endometrial carcinoma & 306498  & 5  \\
Pyruvate Carboxylase Deficiency & PYR & PC & neurometabolic disorder & 3008  & 2  \\
Renpenning & REN & PQBP1 & intellectual disability syndrome & 3242  & 4  \\
Restrictive Dermopathy & RES & ZMPSTE24 & congenital genodermatosis  & 1662  & 5  \\
Rhizomelic Chondrodysplasia Punctata & RHO & PEX7 & group of diseases & 177  & 5  \\
Roberts & ROB & ESCO2 & ? & 3103  & 5  \\
Spinal arteriovenous metameric syndrome & SAM & GSC & ? & 53721  & 1  \\
Sandhoff Disease & SAND & ? & Progressive psychomotor deterioration & 796  & 5  \\
Sialidosis type 2 & SIA & NEU1 & lysosomal storage disease & 87876  & 5  \\
Sickle Cell Anemia & SIC & HBB & Chronic hemolytic anemia & 232  & 3  \\
Proximal Spinal Muscular Atrophy & SPI & SMN1 & neuromuscular disorder & 70  & 4  \\
Spondylodysplastic Ehlers-danlos Syndrome & SPO & ? & Platyspondyly & 536471  & 3  \\
Congenital sucrase-isomaltase deficiency & SUC & SI & carbohydrate intolerance disorder & 35122  & 2  \\
Isolated sulfite oxidase deficiency & SUL & SO & ? & 99731  & 5  \\
Temple syndrome & TEMP & ? & Maturity-onset diabetes of the young & 254516  & 5  \\
Turner Syndrome & TUR & ? & Biliary cirrhosis & 881  & 4  \\
Tyrosinemia Type 1 & TYR & FAH & inborn error of tyrosine catabolism & 882  & 5  \\
Usher Syndrome type 1 & USHB & MYO7A & ? & 231169  & 4  \\
CACH syndrome & VAN & EIF2B5 & ? & 135  & 5  \\
Hyperostosis corticalis generalisata & VANB & ? & craniotubular hyperostosis & 3416  & 2  \\
Walker Warburg & WAL & POMT1 & congenital muscular dystrophy  & 899  & 5  \\
Warsaw Breakage Syndrome & WAR & DDX11 & psychomotor retardation & 280558  & 5  \\
Wolf-hirschhorn Syndrome & WOL & LETM1, NSD2 & Rib segmentation abnormalities & 280  & 5  \\
Zellweger & ZEL & ? & peroxisome biogenesis disorder  & 912  & 5  \\
\end{longtable}

\footnotetext{? indicates that the information is not available.}

\normalsize 

\clearpage
\subsection{Dataset distribution and organization}
\label{app:dataset_org}
\begin{figure}[h]
    \centering
    \begin{subfigure}[t]{0.48\textwidth}
        \centering
        \includegraphics[height=7cm]{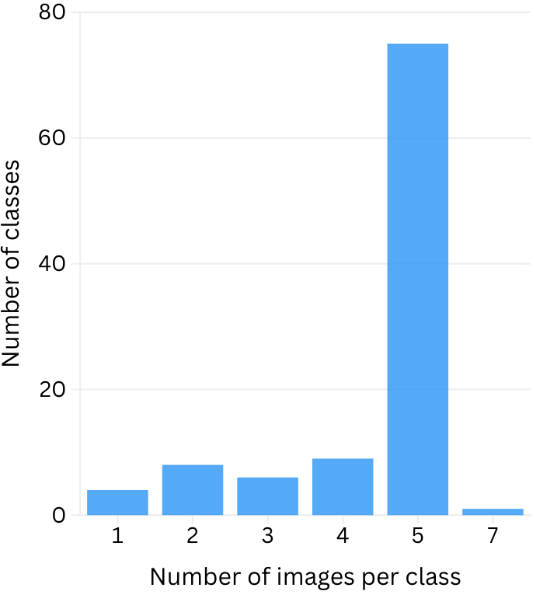}
        \caption{Distribution of images across disease classes.}
        \label{fig:distribution}
    \end{subfigure}
    \hfill
    \begin{subfigure}[t]{0.48\textwidth}
        \centering
        \includegraphics[height=7cm]{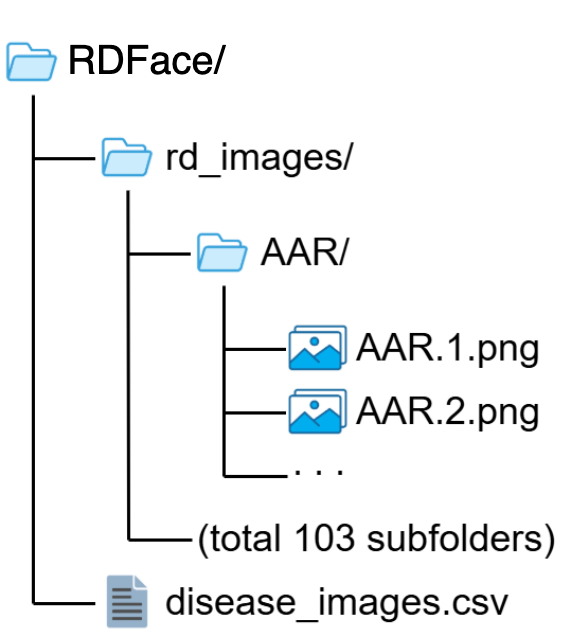}
        \caption{Structure of the RDFace dataset.}
        \label{fig:structure}
    \end{subfigure}
    
    \caption{\textbf{Overview of the RDFace dataset.}}
    \label{fig:s_dataset_overview}
\end{figure}

Aside from the metadata table, we also provide two figures to illustrate the distribution and structure of the RDFace dataset. \Cref{fig:distribution} shows the number of images available for each disease class, highlighting the inherent imbalance in data availability across different rare diseases. \Cref{fig:structure} depicts the organization of the dataset. These visualizations provide a clearer understanding of the dataset's composition and the challenges posed by data scarcity in rare disease research.

\clearpage
\section{Few-shot learning}
\label{app:fewshot_learning}
Few-shot learning (FSL) addresses the problem of classifying instances when only a few labeled examples are available for each class. This setting is common in domains like rare disease diagnosis, where collecting large-scale labeled datasets is impractical due to the low prevalence and data privacy constraints.

In the standard $n$-way $k$-shot FSL setting, each learning episode involves $n$ distinct classes, with only $k$ labeled examples (support set) per class. The goal is to classify unlabeled examples (query set) drawn from the same $n$ classes, using the limited support data for reference.

Formally, each episode consists of:
\begin{itemize}[leftmargin=5em, itemsep=0pt, topsep=0pt]
    \item A support set $\mathcal{S} = \{(x_i, y_i)\}_{i=1}^{n \cdot k}$, containing $k$ samples per class.
    \item A query set $\mathcal{Q} = \{(x_j, y_j)\}_{j=1}^{q}$, containing unseen samples from the same $n$ classes.
\end{itemize}

We adopted Prototypical Networks to solve the few-shot classification task on RDFace. This approach learns an embedding function $f_\theta: \mathcal{X} \rightarrow \mathbb{R}^d$ that maps input images into a feature space. For each class $c_i$, a prototype (centroid) is computed as the mean embedding of its support samples:
\begin{equation}
\mu_i = \frac{1}{k} \sum_{x \in \mathcal{S}_i} f_\theta(x)
\label{eq:prototype}
\end{equation}
For each query image $x^{(q)}$, the model computes the distance between its embedding and each prototype $\mu_i$, typically using the squared Euclidean distance:
\begin{equation}
d(x^{(q)}, \mu_i) = \left\|f_\theta(x^{(q)}) - \mu_i\right\|_2^2
\label{eq:distance}
\end{equation}
The query image is then assigned to the class with the closest prototype, and a cross-entropy loss is computed over the predictions for optimization.

\subsection{Prototypical networks algorithm}
We provide the algorithm used for episodic training on RDFace as below:
\begin{algorithm}[H]
\caption{Prototypical Networks Episodic Training on RDFace}
\label{alg:protonet}
\begin{algorithmic}[1]
\REQUIRE Feature extractor $f_\theta$, training classes $\mathcal{C}_{\text{train}}$, number of ways $n$, support size $k=1$
\STATE Sample $n$ classes $\{c_1, \dots, c_n\} \sim \mathcal{C}_{\text{train}}$
\FOR{each class $c_i$}
    \STATE Sample support set $\mathcal{S}_i = \{x_i^{(s)}\}$, query set $\mathcal{Q}_i = \{x_i^{(q)}\}$
    \STATE Compute prototype: $\mu_i \leftarrow f_\theta(x_i^{(s)})$
\ENDFOR
\FOR{each query $x_j^{(q)} \in \bigcup_i \mathcal{Q}_i$}
    \STATE Compute distances: $d_{ij} \leftarrow \|f_\theta(x_j^{(q)}) - \mu_i\|_2$
    \STATE Predict label: $\hat{y}_j \leftarrow \arg\min_i d_{ij}$
\ENDFOR
\STATE Compute loss: $\mathcal{L}_{\text{episode}} \leftarrow \text{CrossEntropy}(\hat{y}, y)$
\STATE Update $f_\theta$ via backpropagation
\end{algorithmic}
\end{algorithm}

\clearpage
\section{Synthetic image samples}
\label{app:synthetic_images}
\begin{figure}[H]
    \centering
    \includegraphics[width=0.9\textwidth]{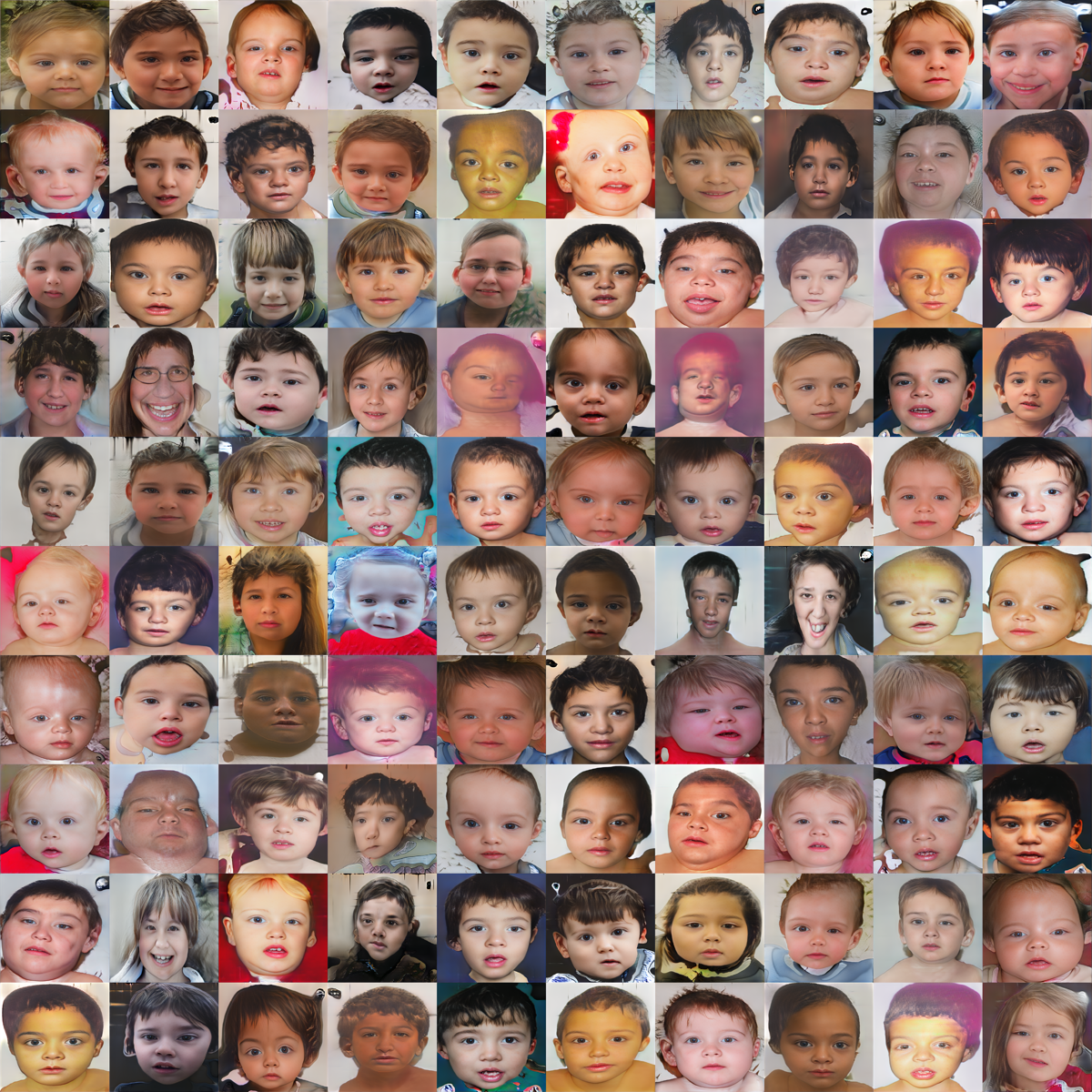}
    \caption{\textbf{Representative FastGAN images ranked most similar to diseases.}}
    \label{fig:fastgan_samples}
\end{figure}

\begin{figure}[H]
    \centering
    \includegraphics[height=\textheight]{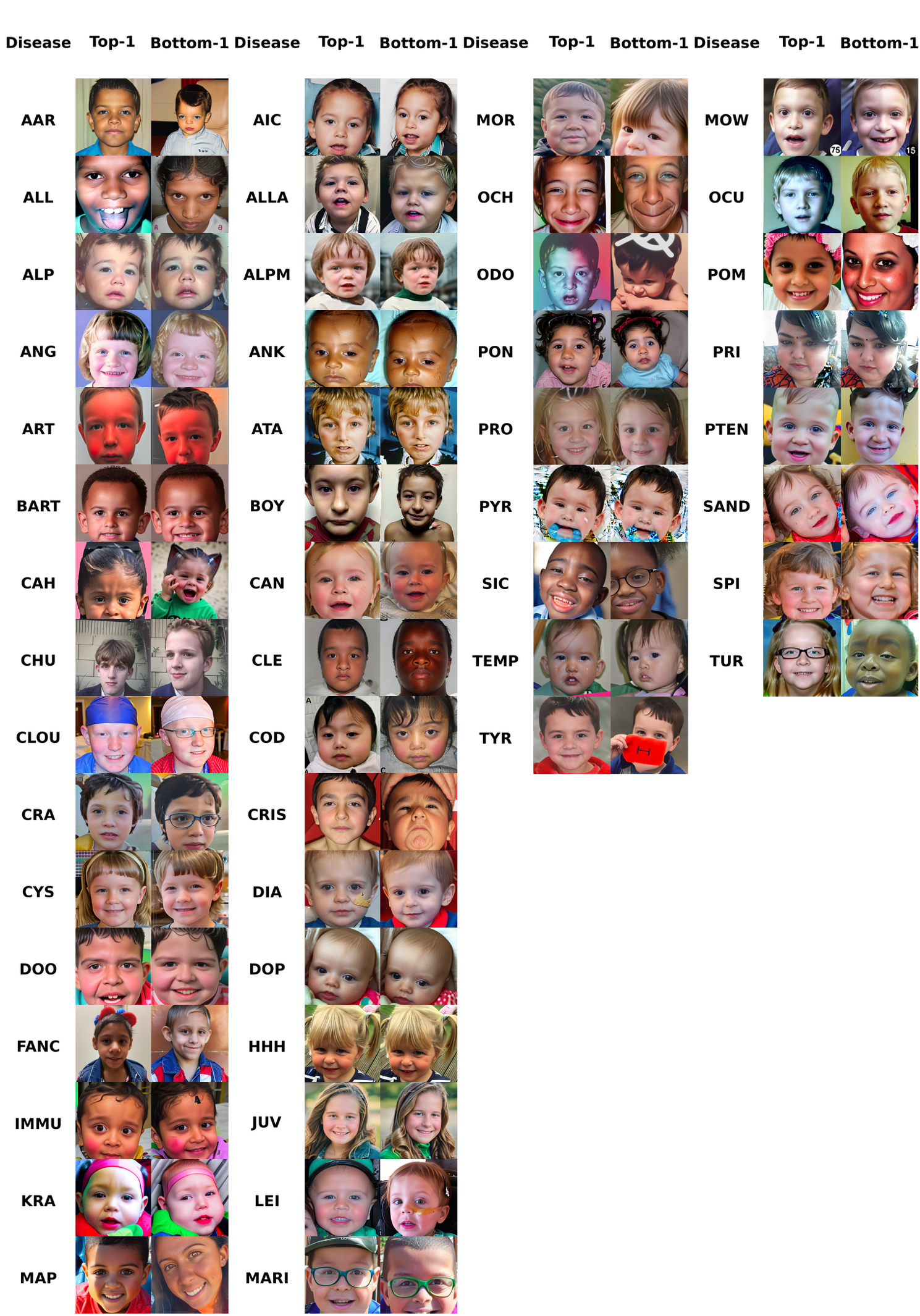}
    \caption{\textbf{Top-1 and Bottom-1 synthetic images based on similarity to real images.} Top-1 and Bottom-1 denote the most and least similar samples respectively.}
    \label{fig:syn_dreambooth}
\end{figure}

\Cref{fig:fastgan_samples} presents the top 100 synthetic facial images generated by FastGAN, ranked by their similarity to real disease faces using a landmark-based cosine metric. These samples highlight the model’s ability to capture coarse facial structure across diverse rare disease classes, although occasional artifacts and inconsistencies remain visible. In contrast, \Cref{fig:syn_dreambooth} provides a more targeted comparison using DreamBooth of 50 classes choosen to display. For each disease class, we show the most and least phenotype-consistent synthetic images compared to real images based on $5 \times 5$ facial landmark cosine similarity. This comparison illustrates both the strength and the limitations of current generative models. While top-ranked DreamBooth samples often replicate key craniofacial traits, the bottom-ranked samples reveal potential risks of phenotype distortion or mode collapse. These visualizations support the use of generative models for phenotype augmentation, while also emphasizing the need for careful validation when applying synthetic images in clinical or diagnostic settings.

\clearpage
\section{Expert and automated evaluation of synthetic data}
\label{app:synthetic_evaluation}

\subsection{Landmark-based similarity analysis}
\label{app:landmark-similarity}
We conducted our evaluation across all 103 rare disease classes in RDFace. For each class, DreamBooth-generated synthetic images were used, provided they passed quality checks based on facial structure and alignment. Images failing these checks, such as those with low RetinaFace detection confidence or invalid landmark configurations, were excluded at the image level, not the class level. All similarity analyses were performed using normalized $5 \times 5$ landmark distance matrices computed from RetinaFace keypoints.

\subsubsection{Heatmap of landmark-based cosine similarities}
\begin{figure}[H]
\centering
\includegraphics[width=0.9\textwidth]{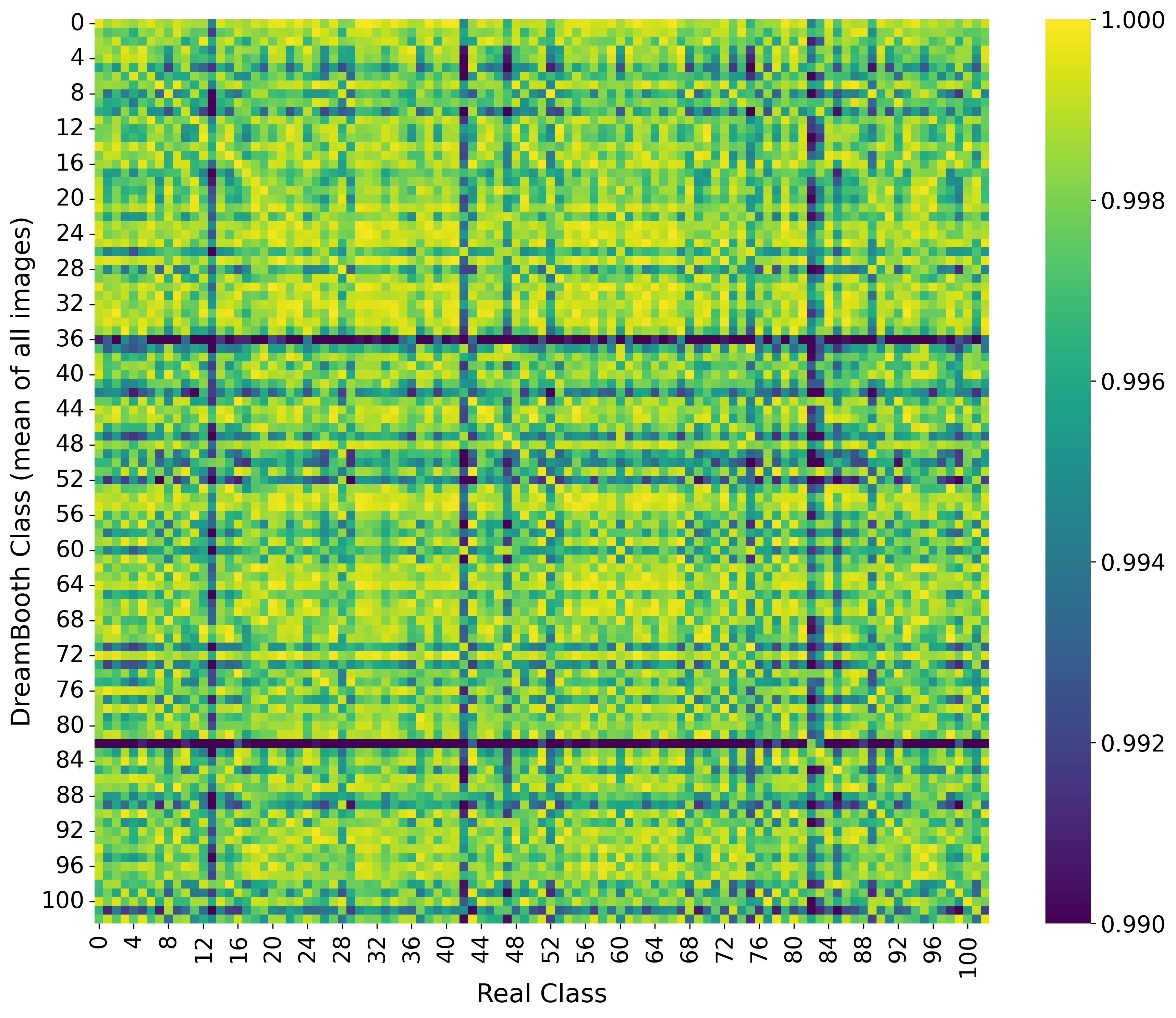}
\caption{\textbf{Cosine similarity heatmap between DreamBooth-generated and real disease prototypes.} The heatmap compares the average landmark distance matrices of DreamBooth-generated images (rows) to those of real disease class prototypes (columns). Brighter values indicate greater similarity.}
\label{fig:dreambooth_heatmap}
\end{figure}

\Cref{fig:dreambooth_heatmap} presents a cosine similarity heatmap comparing the average DreamBooth-generated landmark structure for each class against all real disease class prototypes. The heatmap reveals a strong diagonal trend, reflecting high intra-class similarity, with several classes showing close alignment between synthetic and real facial geometry. However, for certain classes—particularly those with very limited or noisy training data—the diagonal intensity diminishes, indicating lower phenotype fidelity. Off-diagonal activity highlights cross-class resemblance, often reflecting overlapping craniofacial features among syndromes or mode collapse in synthesis.

\subsubsection{Ranking consistency of DreamBooth-generated Images}
\begin{figure}[H]
\centering
\includegraphics[width=0.8\textwidth]{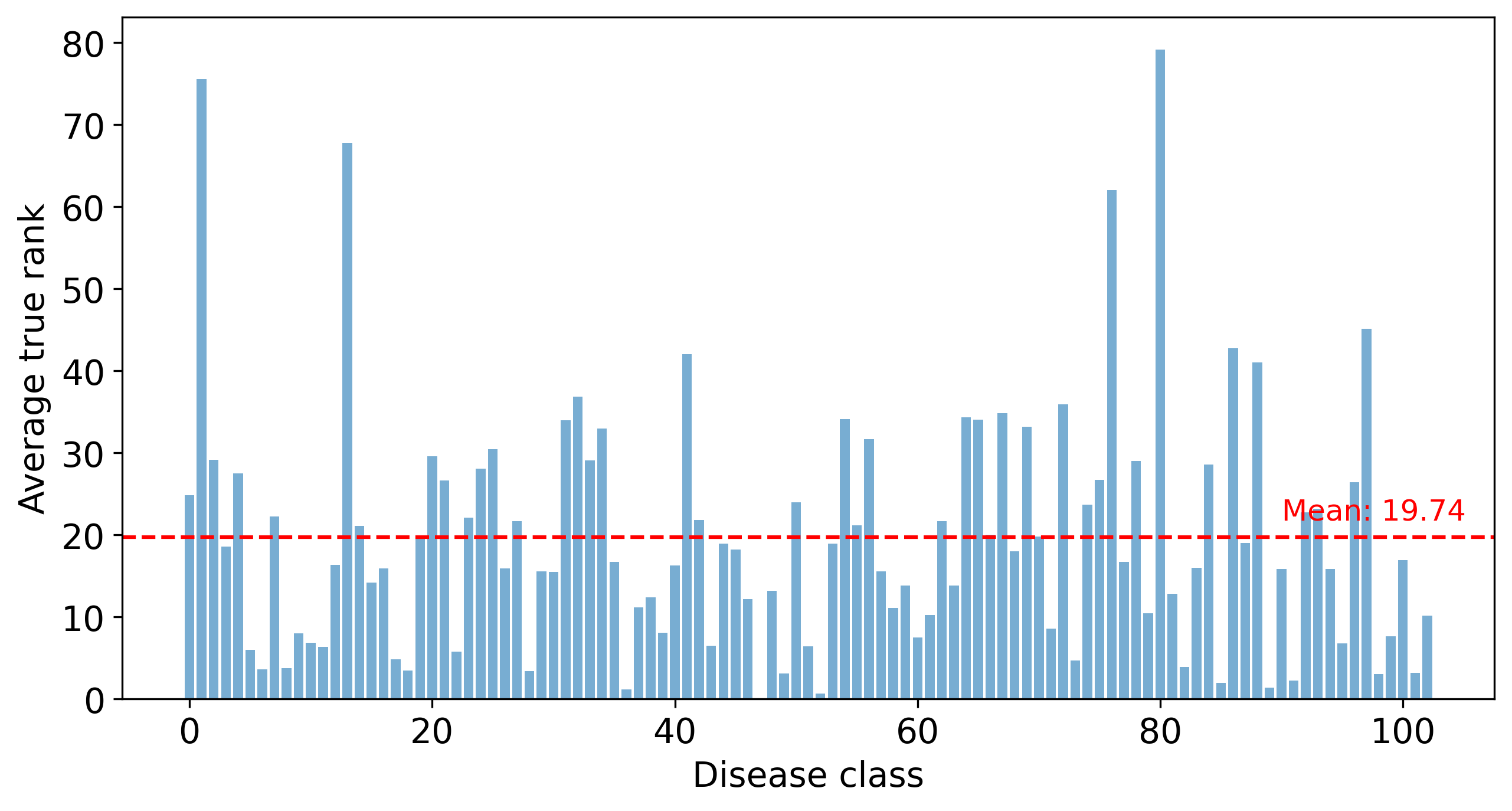}
\caption{\textbf{Average rank of the true disease class for each DreamBooth-generated image.} Each rank is calculated using cosine similarity to the corresponding real class prototype. The red dashed line indicates the overall mean rank.}
\label{fig:average_rank}
\end{figure}

To quantify how consistently DreamBooth preserves class-specific features, we computed the rank position of the ground-truth class for each synthetic image among all real class prototypes based on cosine similarity. \Cref{fig:average_rank} summarizes the average rank per class.

While many classes exhibit ranks close to 1, several fall well below the mean, suggesting inconsistency in capturing distinctive morphology. These cases often correspond to underrepresented or visually subtle conditions in the training set. The overall mean rank across all classes is 19.74, serving as a practical reference for DreamBooth’s phenotype alignment under ultra-low-shot conditions.

\subsection{Expert review}
\label{app:expert_review}
In order to evaluate the clinical plausibility of synthetic image-label pairs, we conducted an expert review of 50 DreamBooth and 50 FastGAN samples. Each image was independently assessed by two medical doctor (MD) students (MD1 and MD2) and categorized as \textit{Plausible}, \textit{Implausible}, or \textit{Uncertain}. The results of this expert evaluation are summarized in \Cref{tab:synthetic_label_validation}.

\begin{table}[h]
\centering
\caption{Expert evaluation of the plausibility of synthetic image-label pairs generated by DreamBooth and FastGAN with two MDs.}
\begin{tabular}{l|cccc}
\toprule
\textbf{Plausibility Rating} & \multicolumn{2}{c}{\textbf{DreamBooth}} & \multicolumn{2}{c}{\textbf{FastGAN}} \\
\cmidrule(lr){2-3} \cmidrule(lr){4-5}
 & MD1 & MD2 & MD1 & MD2 \\
\midrule
Plausible      & 38 & 31 & 14 & 5 \\
Implausible    & 2  & 3  & 15 & 30 \\
Uncertain      & 10 & 16 & 21 & 15 \\
\bottomrule
\end{tabular}
\label{tab:synthetic_label_validation}
\end{table}

To assess the consistency between the two raters, inter-rater reliability was computed for each model, and the pairwise confusion matrices are presented in \Cref{tab:synthetic_label_inter_rate}. Each cell represents the number of images assigned to a given label combination by the two MD reviewers (\textit{DreamBooth / FastGAN}). A strong concentration of counts along the diagonal indicates higher agreement on image plausibility.

\begin{table}[h]
\centering
\caption{Expert validation confusion matrices for DreamBooth (DB) and FastGAN (FG). (DB / FG)}
\vspace{0.3em}
\renewcommand{\arraystretch}{1.2}
\setlength{\tabcolsep}{6pt}
\begin{tabular}{lccc|c}
\toprule
\textbf{Label} & \textbf{Plausible} & \textbf{Implausible} & \textbf{Uncertain} & \textbf{Total} \\
\midrule
\textbf{Plausible}   & 31\,/\,2  & 0\,/\,0  & 0\,/\,3  & 31\,/\,5  \\
\textbf{Implausible} & 0\,/\,7   & 2\,/\,11 & 1\,/\,12 & 3\,/\,30 \\
\textbf{Uncertain}   & 7\,/\,5   & 0\,/\,4  & 9\,/\,6  & 16\,/\,15 \\
\midrule
\textbf{Total}       & 38\,/\,14 & 2\,/\,15 & 10\,/\,21 & 50\,/\,50 \\
\bottomrule
\end{tabular}
\label{tab:synthetic_label_inter_rate}
\end{table}

Quantitatively, the inter-rater agreement results (\Cref{tab:synthetic_label_inter_rates_stat}) demonstrate that DreamBooth achieved a substantially higher degree of consensus between raters (\(\kappa = 0.654\)), corresponding to ``substantial agreement'' on the Landis–Koch scale, whereas FastGAN showed only minimal agreement (\(\kappa = 0.069\)). This difference highlights that DreamBooth-generated samples were generally perceived as more clinically plausible and consistent across evaluators, while FastGAN outputs exhibited greater variability and uncertainty.

\begin{table}[h]
\centering
\caption{Inter-rater agreement metrics for DreamBooth (DB) and FastGAN (FG).}
\vspace{0.3em}
\renewcommand{\arraystretch}{1.2}
\setlength{\tabcolsep}{8pt}
\begin{tabular}{l|cccc}
\toprule
\textbf{Method} & \textbf{Observed Agreement (\%)} & \textbf{Cohen’s $\kappa$} & \textbf{SE} & \textbf{95\% CI} \\
\midrule
DB & 84.0 (42/50) & 0.654 & 0.106 & [0.446, 0.862] \\
FG & 38.0 (19/50) & 0.069 & 0.091 & [--0.110, 0.248] \\
\bottomrule
\end{tabular}
\label{tab:synthetic_label_inter_rates_stat}
\end{table}

Overall, the expert review confirms that DreamBooth produces synthetic facial images that retain phenotypic plausibility and diagnostic relevance more consistently than FastGAN, aligning with the quantitative similarity metrics and qualitative visual inspection presented in the main manuscript.

\subsection{Observations and implications}
Our evaluation highlights the complementary strengths of automated and expert-based assessments for characterizing the quality of synthetic facial images. Landmark-based similarity analysis reveals that many disease classes exhibit distinct and consistent craniofacial geometry, validating the use of facial landmarks as a phenotypic signature. The observed variability in average rank and cosine similarity across classes reflects sensitivity to both morphological subtlety and the quality of training exemplars. These findings indicate that normalized landmark distance metrics offer an interpretable, spatially grounded method for quantifying phenotype preservation and filtering synthetic samples prior to downstream tasks.

However, structural alignment alone does not guarantee clinical plausibility. To address this, we conducted an expert review to evaluate whether synthetic image-label pairs appeared medically credible. Results show that DreamBooth-generated images were more frequently judged as plausible by at least one medical doctor, whereas FastGAN samples showed higher rates of uncertainty and implausibility. These findings underscore that perceptual realism, which is often emphasized in generative model benchmarks, does not always correlate with diagnostic fidelity. Expert review provides a critical semantic layer that captures domain-specific context often missed by automated metrics.

Together, these results suggest that effective synthetic data curation requires a multi-faceted evaluation strategy that integrates spatial similarity, visual quality, and human judgment. Such approaches are especially important in rare disease settings, where subtle phenotypic cues and label noise can dramatically impact model performance. In future applications, combining interpretable landmark filtering with lightweight expert-in-the-loop review may provide a scalable path for enhancing dataset quality and clinical utility.

\clearpage
\section{Tradeoff between disease-specific structure and visual realism}
\label{app:synthetic-metrics}

To assess the quality and disease-relevance of generated synthetic images, we evaluate two image-level realism metrics: RetinaFace detection confidence and LPIPS perceptual similarity. These metrics are computed across Top-$n$ subsets (ranging from 1000 to 6000 images), where the images are ranked by their landmark-based cosine similarity to real disease class prototypes.

\begin{figure}[h]
    \centering
    \includegraphics[width=0.9\textwidth]{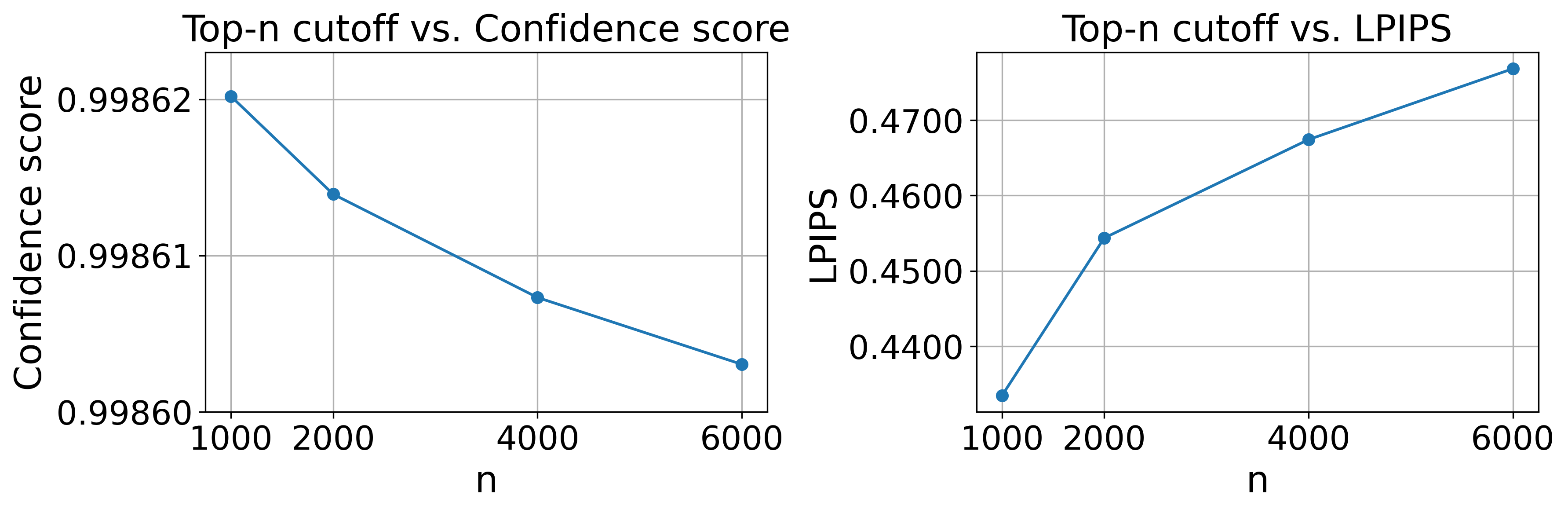}
    \caption{\textbf{DreamBooth – Correlation Between Top-$n$ Ranking and Visual Realism.} 
    RetinaFace detection confidence (left) and LPIPS similarity (right) across Top-$n$ ranked DreamBooth images.}
    \label{fig:s3_relationship}
\end{figure}

\paragraph{DreamBooth}
\Cref{fig:s3_relationship} shows the trend of RetinaFace and LPIPS scores across Top-$n$ DreamBooth subsets. As $n$ increases, RetinaFace confidence scores gradually decline, suggesting reduced alignment and detectability in lower-ranked images. Meanwhile, LPIPS scores increase, indicating a decrease in perceptual similarity to real images. These opposing trends reflect a trade-off between structural fidelity (as captured by landmarks) and low-level texture realism. The landmark-based ranking effectively promotes DreamBooth samples with coherent facial geometry and higher visual plausibility.

\begin{figure}[h]
    \centering
    \includegraphics[width=0.9\textwidth]{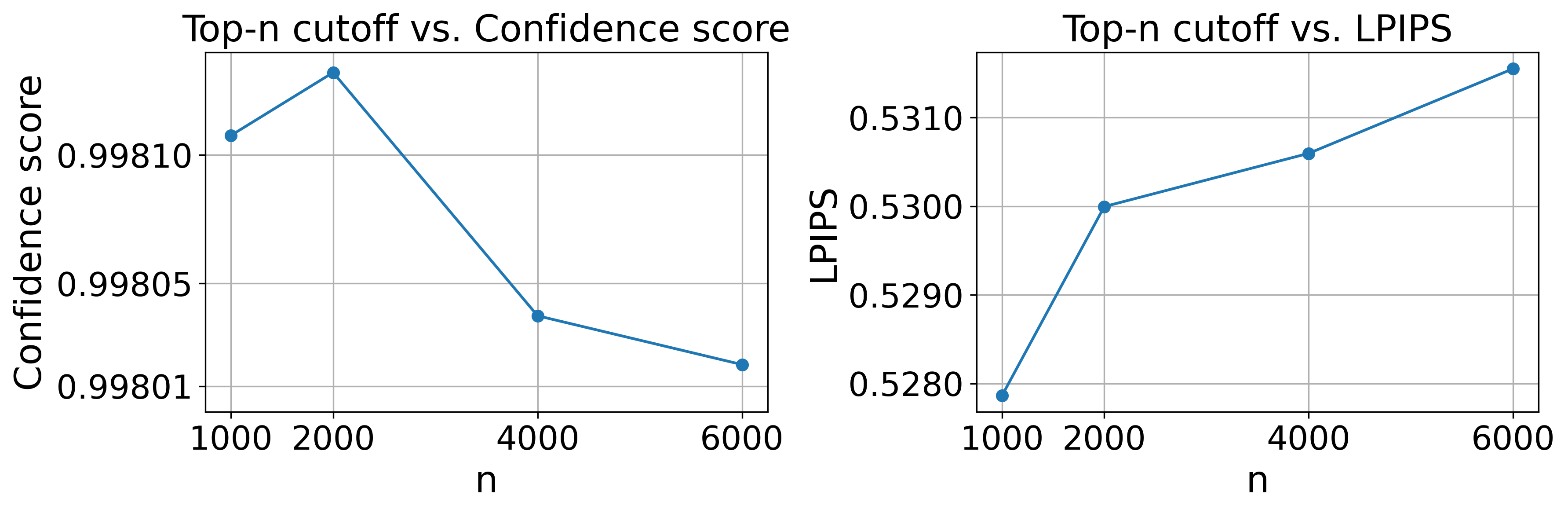}
    \caption{\textbf{FastGAN – Correlation Between Top-$n$ Ranking and Visual Realism.} 
    RetinaFace detection confidence (left) and LPIPS similarity (right) across Top-$n$ ranked FastGAN images.}
    \label{fig:s4_relationship}
\end{figure}

\paragraph{FastGAN}
FastGAN samples exhibit a broadly similar trend to DreamBooth in terms of ranking-based visual quality (see \Cref{fig:s4_relationship}). As the Top-$n$ threshold increases, RetinaFace confidence scores slightly decline and LPIPS scores gradually increase, indicating reduced structural detectability and perceptual similarity at lower-ranked positions. However, some local fluctuations are observed—particularly around the Top-2000 cutoff—where both metrics deviate slightly from the overall trajectory. These irregularities suggest that landmark-based ranking is still somewhat effective for FastGAN, but may be less stable than for DreamBooth due to the lack of class-specific conditioning.

\clearpage
\section{Synthetic data-involved downstream tasks results}
\subsection{Standard supervised classification and synthetic scaling effect}
\label{app:downstream_results_classification}
\Cref{tab:topk_aug_dreambooth} and \Cref{tab:topk_aug_fastgan} below report Top-$k$ classification accuracies across various backbones. Each row represents classification performance (Top-$k$ accuracy) under a different training configuration. “Real only” refers to models trained exclusively on real RDFace data. “Top-$n$” rows correspond to training sets augmented with the Top-$n$ synthetic images selected based on landmark similarity to real samples. The synthetic images are chosen to best align with phenotype-specific facial structure.

\renewcommand{\arraystretch}{0.9} 

\setlength{\tabcolsep}{1pt}
\begin{table}[H]
\centering
\caption{Top-$k$ accuracies (\%) across backbones and synthetic cutoffs of DreamBooth samples.}
\label{tab:topk_aug_dreambooth}
\small
\begin{tabularx}{\textwidth}{l*{7}{>{\centering\arraybackslash}X}}

\toprule
\textbf{ACC (\%)} & \textbf{Top-$n$} & \textbf{ResNet} & \textbf{DenseNet} & \textbf{FaceNet} & \textbf{VGG} & \textbf{Swin-T} & \textbf{CLIP} \\
\midrule

\multirow{5}{*}{Top-1}
    & Real only         &   6.90 (1.45)     &    \textbf{15.93 (2.34)}    &    9.91 (1.81)     &   11.68 (1.58)    &    14.34 (2.61)    &    3.1 (1.48)     \\
    & Top-1000   &   12.21 (1.70)    &    \textbf{17.52 (2.29)}    &    15.04 (2.87)    &   16.64 (4.07)    &    16.81 (1.40)    &    9.03 (2.29)    \\
    & Top-2000   &   12.57 (2.61)    &    \textbf{20.35 (1.40)}    &    14.51 (1.34)    &   14.69 (1.61)    &    18.76 (2.20)    &    15.22 (4.35)    \\
    & Top-4000   &   13.27 (2.26)    &    \textbf{19.65 (0.74)}    &    16.46 (1.01)    &   17.35 (4.18)    &    18.76 (2.29)   &    16.81 (1.25)   \\
    & Top-6000   &   13.63 (1.61)    &    \textbf{21.06 (2.53)}    &    16.28 (2.91)    &   16.64 (2.61)    &    18.94 (2.70)   &    15.75 (2.58)   \\
\midrule

\multirow{5}{*}{Top-5}
    & Real only         &   18.58 (3.00)    &    \textbf{33.63 (3.70)}    &    24.60 (5.43)    &    29.91 (2.68)    &    26.19 (2.68)    &   12.74 (2.84)    \\
    & Top-1000   &   27.26 (3.39)    &    \textbf{35.58 (3.39)}    &    32.04 (1.92)    &    29.91 (1.45)    &    33.81 (2.29)    &   17.52 (4.26)    \\
    & Top-2000   &   30.09 (1.98)    &    \textbf{37.35 (2.37)}    &    33.63 (2.65)    &    33.10 (3.99)    &    36.81 (3.40)    &   25.13 (2.70)    \\
    & Top-4000   &   30.09 (4.29)    &    \textbf{40.35 (4.08)}    &    33.98 (2.97)    &    34.16 (2.84)    &    35.40 (4.15)    &   28.14 (0.97)    \\
    & Top-6000   &   33.45 (4.70)    &     \textbf{39.65 (2.29)}   &    32.74 (1.77)    &    33.81 (1.58)    &    36.46 (3.78)    &   29.38 (2.45)    \\
\midrule

\multirow{5}{*}{Top-10}
    & Real only         &   28.50 (3.67)    &    \textbf{43.01 (2.63)}    &    34.87 (4.75)    &    38.41 (1.34)    &    35.93 (3.17)    &   19.12 (5.18)    \\
    & Top-1000   &   38.41 (3.94)    &    \textbf{47.61 (3.30)}    &    44.25 (2.94)    &    39.82 (1.77)    &    46.19 (3.03)    &   26.02 (2.55)    \\
    & Top-2000   &   41.59 (5.16)    &    \textbf{47.26 (2.13)}    &    43.89 (2.55)    &    43.01 (1.34)    &    46.02 (1.08)    &   33.98 (2.25)    \\
    & Top-4000   &   40.53 (3.45)    &    \textbf{51.33 (3.00)}    &    43.72 (2.70)    &    44.25 (3.00)    &    46.19 (1.58)    &   34.87 (1.61)    \\
    & Top-6000   &   42.65 (2.68)    &   \textbf{49.20 (1.48)}     &    43.89 (2.47)    &    46.73 (1.15)    &    48.67 (2.26)    &   37.70 (2.03)    \\
\midrule

\multirow{5}{*}{Top-30}
    & Real only         &   54.34 (2.39)    &    \textbf{64.42 (1.92)}    &    58.23 (5.06)    &    60.88 (2.02)    &    58.41 (3.81)    &   42.30 (4.40)    \\
    & Top-1000   &   62.30 (2.22)    &   \textbf{ 70.62 (3.03)}    &    64.07 (3.99)    &    63.01 (2.83)    &    \textbf{70.62 (2.20)}    &   49.56 (2.65)    \\
    & Top-2000   &   62.12 (3.15)    &    \textbf{70.62 (3.39)}    &    64.60 (3.59)    &    66.73 (3.99)    &    69.56 (2.70)    &   57.17 (4.18)    \\
    & Top-4000   &   63.36 (2.91)    &   70.09 (1.48)     &    67.96 (1.58)    &    64.42 (2.37)    &    \textbf{70.80 (1.08)}    &   57.35 (2.02)    \\
    & Top-6000   &   63.19 (3.94)    &    \textbf{68.67 (3.23)}    &    63.01 (2.89)    &    66.37 (2.58)    &    68.50 (2.39)    &   61.24 (3.05)    \\
\bottomrule
\end{tabularx}
\end{table}

\setlength{\tabcolsep}{1pt}
\begin{table}[H]
\centering
\caption{Top-$k$ accuracies (\%) across backbones and synthetic cutoffs of FastGAN samples.}
\label{tab:topk_aug_fastgan}
\small
\begin{tabularx}{\textwidth}{l*{7}{>{\centering\arraybackslash}X}}
\toprule
\textbf{ACC (\%)} & \textbf{Top-$n$} & \textbf{ResNet} & \textbf{DenseNet} & \textbf{FaceNet} & \textbf{VGG} & \textbf{Swin-T} & \textbf{CLIP} \\
\midrule

\multirow{5}{*}{Top-1}
    & Real only         &   6.90 (1.45)     &    \textbf{15.93 (2.34)}    &    9.91 (1.81)     &   11.68 (1.58)    &    14.34 (2.61)    &    3.1 (1.48)     \\
    & Top-1000   &   8.50 (1.48)    &    \textbf{13.27 (3.13)}    &    6.55 (2.84)    &    7.26 (2.37)    &    10.44 (1.70)    &    1.42 (1.34)   \\
    & Top-2000   &   6.55 (2.13)    &    \textbf{10.44 (0.40)}    &    5.13 (0.74)    &    6.37 (3.67)    &    8.50 (2.63)     &    1.06 (1.15)   \\
    & Top-4000   &   4.07 (1.48)    &    8.14 (1.45)     &    4.96 (1.19)    &    7.96 (2.26)    &    \textbf{9.73 (1.40)}     &    0.71 (0.74)   \\
    & Top-6000   &   4.78 (0.79)    &    \textbf{9.73 (2.08)}     &    4.78 (1.34)    &    5.49 (2.61)    &    9.20 (1.84)     &    1.42 (1.01)   \\
\midrule

\multirow{5}{*}{Top-5}
    & Real only         &   18.58 (3.00)    &    \textbf{33.63 (3.70)}    &    24.60 (5.43)    &    29.91 (2.68)    &    26.19 (2.68)    &   12.74 (2.84)    \\
    & Top-1000   &   20.71 (1.34)    &    \textbf{27.79 (1.34)}    &    18.05 (2.97)    &     18.94 (3.40)   &    25.66 (1.88)    &   5.49 (2.37)    \\
    & Top-2000   &   19.12 (2.31)    &    23.54 (2.22)    &    18.05 (1.73)    &     18.76 (2.89)   &   \textbf{25.49 (4.03)}    &   6.90 (2.76)    \\
    & Top-4000   &   17.17 (2.31)    &    25.49 (2.89)    &    16.81 (1.25)    &     20.71 (2.04)   &    \textbf{26.19 (3.57)}    &   5.49 (1.70)    \\
    & Top-6000   &   18.58 (1.98)    &    \textbf{25.49 (2.02)}    &    15.58 (1.94)    &     21.95 (0.97)   &    23.89 (4.42)    &   6.19 (1.08)    \\
\midrule

\multirow{5}{*}{Top-10}
    & Real only         &   28.50 (3.67)    &    \textbf{43.01 (2.63)}    &    34.87 (4.75)    &    38.41 (1.34)    &    35.93 (3.17)    &   19.12 (5.18)    \\
    & Top-1000   &   29.20 (3.59)    &    \textbf{37.70 (3.29)}    &    30.27 (3.51)    &    29.03 (5.92)    &    36.64 (2.22)    &   11.15 (1.48)   \\
    & Top-2000   &   28.67 (4.18)    &    34.34 (0.74)    &    28.85 (3.79)    &    29.38 (2.20)    &    \textbf{35.58 (3.51)}    &   11.68 (2.20)   \\
    & Top-4000   &   29.03 (2.68)    &    35.93 (3.04)    &    28.32 (2.50)    &    31.33 (4.23)    &    \textbf{36.64 (4.58)}    &   11.15 (2.77)   \\
    & Top-6000   &   31.15 (3.15)    &    \textbf{35.93 (3.10)}    &    24.07 (2.29)    &    32.39 (2.97)    &    33.45 (3.62)    &   13.63 (0.79)   \\
\midrule

\multirow{5}{*}{Top-30}
    & Real only         &   54.34 (2.39)    &    \textbf{64.42 (1.92)}    &    58.23 (5.06)    &    60.88 (2.02)    &    58.41 (3.81)    &   42.30 (4.40)    \\
    & Top-1000   &   51.68 (2.04)    &    \textbf{61.59 (5.82)}    &    57.52 (0.63)    &     55.22 (5.07)   &    60.35 (1.92)    &    34.51 (2.73)   \\
    & Top-2000   &   53.10 (3.54)    &    58.41 (2.08)    &    55.04 (4.44)    &     53.27 (3.33)   &    \textbf{61.77 (4.90)}    &    32.39 (2.13)  \\
    & Top-4000   &   56.46 (3.51)    &    \textbf{60.00 (3.15)}    &    55.58 (1.70)    &     59.12 (4.12)   &    57.88 (3.40)    &    32.57 (2.46)   \\
    & Top-6000   &   55.75 (3.19)    &    54.34 (3.68)    &    55.40 (2.31)    &     \textbf{60.71 (1.94)}   &    56.46 (1.15)    &    38.23 (5.78)   \\
\bottomrule
\end{tabularx}
\end{table}

\begin{figure}[H]
    \centering
    \includegraphics[width=0.83\textwidth]{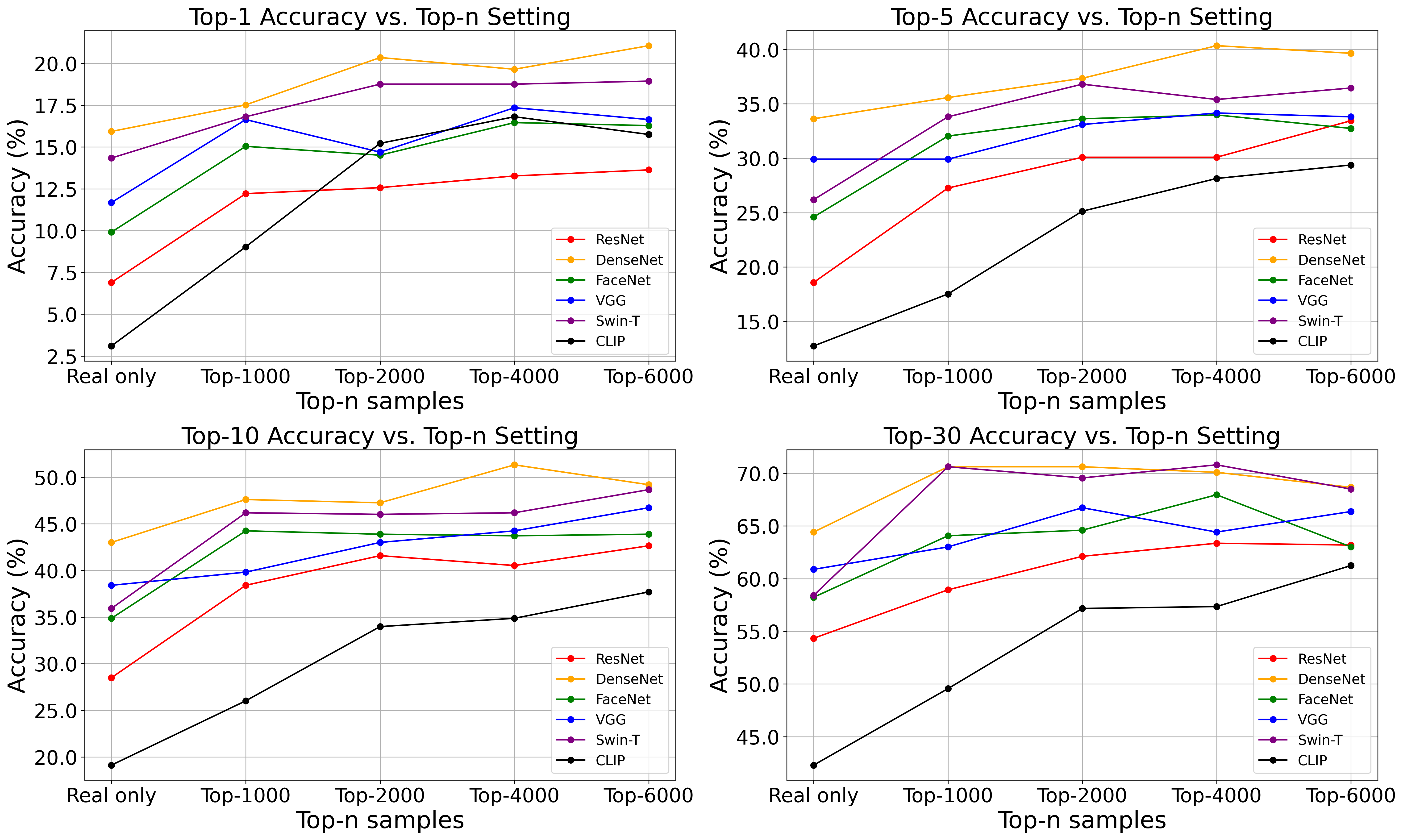}
    \caption{\textbf{Top-$k$ accuracy comparison using DreamBooth-generated data.} Each subplot shows Top-1, Top-5, Top-10, and Top-30 accuracy across synthetic cutoffs for six backbone models. DreamBooth augmentation improves performance across most settings.}
    \label{fig:downstream_db_v}
\end{figure}

\begin{figure}[H]
    \centering
    \includegraphics[width=0.83\textwidth]{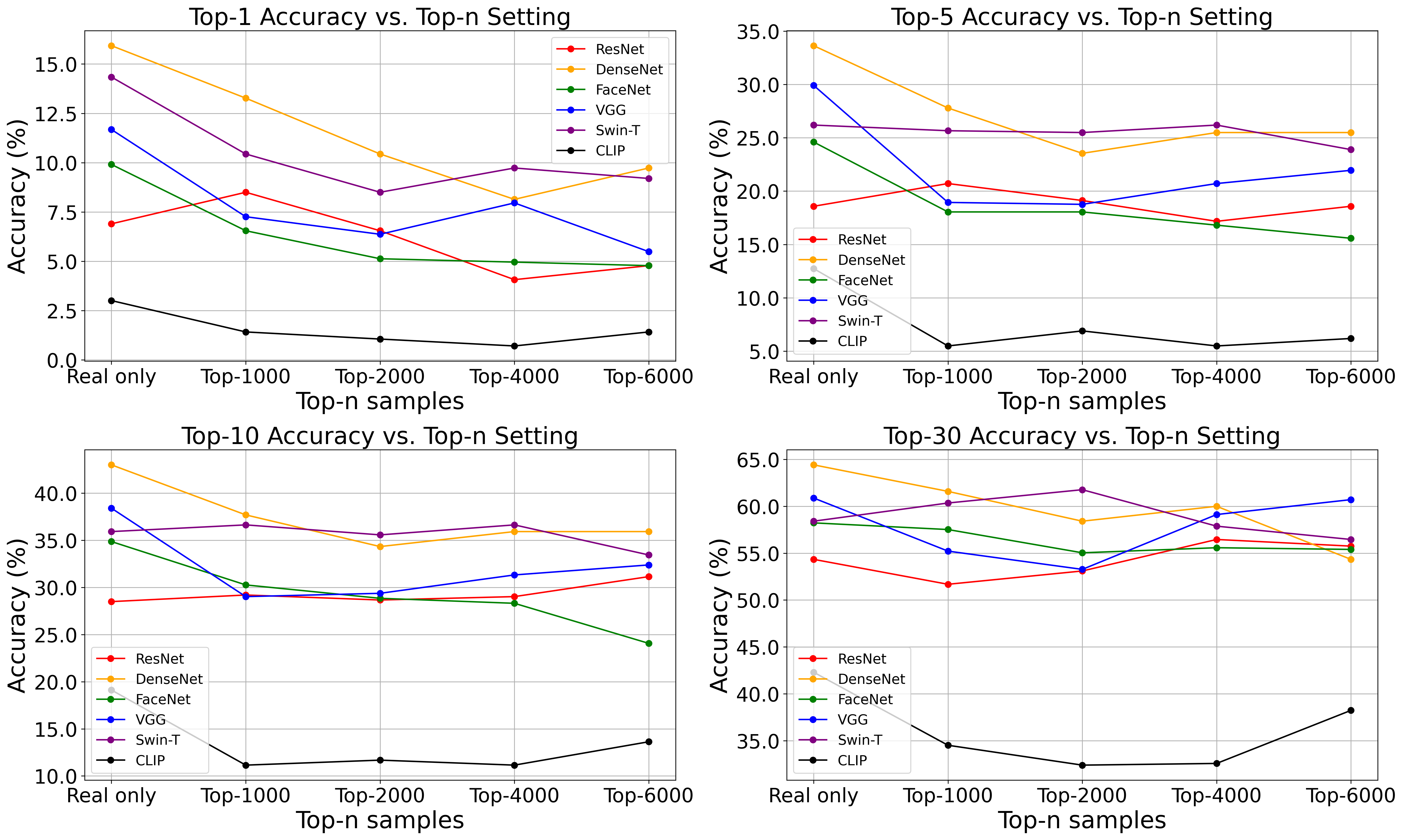}
    \caption{\textbf{Top-$k$ accuracy comparison using FastGAN-generated data.} Each subplot shows Top-1, Top-5, Top-10, and Top-30 accuracy across synthetic cutoffs for six backbone models. Compared to DreamBooth, FastGAN augmentation results in less consistent or degraded performance across most settings.}
    \label{fig:downstream_fg_v}
\end{figure}

\paragraph{Synthetic scaling effect} The relationship between real-only and synthetic-augmented performance across Top-$k$ settings and Top-$n$ synthetic cutoffs is shown in \Cref{fig:downstream_db_v} and \Cref{fig:downstream_fg_v}. These plots visualize the downstream classification accuracies using DreamBooth and FastGAN generated samples, respectively, across six backbone models.

\subsection{Few-shot learning}
\label{app:downstream_results_few_shot}
Based on prior results, we focus our few-shot learning analysis on DreamBooth-augmented data. This choice is motivated by the consistent advantages of DreamBooth samples over FastGAN in terms of semantic fidelity, preservation of disease-relevant facial traits, and interpretability. These properties are especially important in ultra-low-shot settings, where maximizing the realism and diagnostic relevance of synthetic data is critical for generalization.
\Cref{tab:fewshot_aug_results} summarizes performance across multiple backbones under synthetic data augmentation. Each pair of rows shares the same pretrained backbone, where the first row (e.g., ResNet) uses only real training data, and the second row (e.g., ResNet (dream)) incorporates DreamBooth-generated synthetic images for data augmentation.

\setlength{\tabcolsep}{1pt}
\begin{table}[h]
\centering
\caption{Few-shot classification accuracies (\%) under 5-way, 10-way, and 15-way settings with 1-shot and 5-shot support and query sets.}
\label{tab:fewshot_aug_results}
\begin{tabularx}{\textwidth}{l*{6}{>{\centering\arraybackslash}X}}
\toprule
\textbf{ACC (\%)} & \multicolumn{2}{c}{\textbf{5-way}} & \multicolumn{2}{c}{\textbf{10-way}} & \multicolumn{2}{c}{\textbf{15-way}} \\
\cmidrule(lr){2-3} \cmidrule(lr){4-5} \cmidrule(lr){6-7}
& 1-shot & 5-shot & 1-shot & 5-shot & 1-shot & 5-shot \\
\midrule
ResNet          & 24.18 (2.56) & --     & 18.15 (2.58) & --     & 17.99 (1.26) & --     \\
ResNet (dream)      & \textbf{25.72 (1.62)} & 33.62 (1.91) & \textbf{22.21 (2.60)} & 22.63 (2.11) & \textbf{18.22 (1.91)} & 20.04 (2.77) \\
\cmidrule(lr){1-7}
DenseNet          & 26.20 (2.01) & --     & 17.36 (1.66) & --     & \textbf{17.30 (3.45)} & --     \\
DenseNet (dream)     & \textbf{29.88 (1.51)} & 33.40 (2.02) & \textbf{19.79 (3.19)} & 24.43 (3.82) & 16.06 (2.77) & 18.35 (2.24) \\
\cmidrule(lr){1-7}
FaceNet      & \textbf{25.16 (4.89)} & --     & 12.93 (2.33) & --     &  8.03 (1.40) & --     \\
FaceNet (dream)    & 23.60 (4.06) & 28.30 (2.92) & \textbf{13.17 (2.33)} & 17.31 (3.47) & \textbf{ 9.07 (1.44)} & 12.17 (1.24) \\
\cmidrule(lr){1-7}
VGG    & \textbf{21.54 (3.72)} & --     & 10.82 (2.14) & --     &  9.05 (2.20) & --     \\
VGG (dream)    & 21.02 (5.85) & 26.76 (3.42) & \textbf{13.67 (2.08)} & 13.68 (1.94) &  \textbf{9.59 (1.48)} & 11.20 (1.59) \\
\cmidrule(lr){1-7}
Swin-T      & 22.24 (2.88) & --     & 10.94 (2.69) & --     &  8.03 (2.41) & --     \\
Swin-T (dream)  & \textbf{26.72 (4.34)} & 25.78 (4.91) & \textbf{13.30 (1.56)} & 13.43 (1.94) &  \textbf{9.57 (2.57)} &  8.82 (0.97) \\
\cmidrule(lr){1-7}
CLIP         & \textbf{23.48 (5.28)} & --     & \textbf{12.33 (1.96)} & --     &  7.30 (1.90) & --     \\
CLIP (dream)   & 22.30 (2.63) & 24.80 (3.48) & 12.14 (1.98) & 13.04 (2.08) &  \textbf{8.66 (2.82)} &  8.11 (1.32) \\
\bottomrule
\end{tabularx}
\end{table}

\subsection{Observations}

\paragraph{Standard supervised classification} 
The results in \Cref{tab:topk_aug_dreambooth} and \Cref{tab:topk_aug_fastgan} highlight a clear distinction in the effectiveness of different generative models for data augmentation in classifications. DreamBooth consistently enhances model performance across Top-$k$ accuracies and Top-$n$ settings, demonstrating its ability to produce high-quality, semantically aligned images that support generalization. In contrast, FastGAN shows less consistent gains and, in some cases, degrades performance as more synthetic data is added—indicating a lower signal-to-noise ratio in the generated samples. These findings underscore the importance of not only using synthetic data, but selecting the right generation method.

\paragraph{Few-shot learning}
\Cref{tab:fewshot_aug_results} further illustrates how the utility of synthetic augmentation depends on the interaction between model architecture and data quality. While DreamBooth-generated samples consistently improve performance, gains vary across backbones, with CNNs such as DenseNet and ResNet benefiting more than transformer-based models like Swin Transformer or CLIP. Additionally, the performance gap between 1-shot and 5-shot conditions emphasizes the value of even modest increases in labeled support. These results suggest that few-shot learning with synthetic data is most effective when both model and augmentation strategy are jointly optimized for the task.

Overall, high semantic fidelity, structural consistency, and disease-relevant visual features appear essential for synthetic augmentation to benefit rare disease classification, especially in ultra-low-shot regimes.

\clearpage
\section{VLM-based report generation}
\subsection{Prompt design}
\label{app:vlm_prompt}

\begin{figure}[H]
    \centering
    \includegraphics[width=0.67\linewidth]{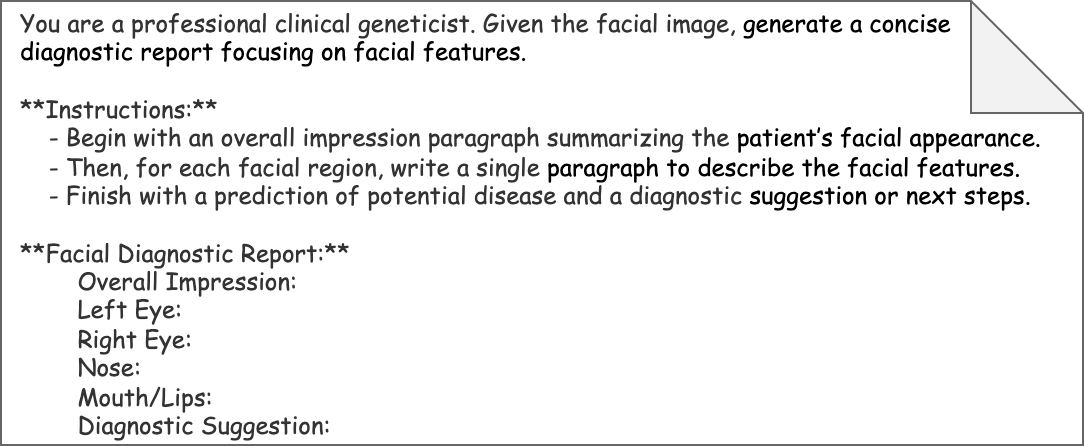}
    \caption{\textbf{Prompt template for VLM-based report generation.}}
    \label{fig:vlm_prompt}
\end{figure}

To ensure consistency and clinical interpretability of the generated phenotype reports, we designed a structured prompt tailored to the rare disease diagnosis task. The prompt guides the multimodal vision language model (VLM) to assume the role of a professional clinical geneticist and produce structured, concise diagnostic reports based on input facial images. The prompt emphasizes anatomical coverage, medical reasoning, and alignment with the supervised landmark annotations used in other parts of our study. A prompt template is shown in \Cref{fig:vlm_prompt}.

\subsection{Report evaluation}
\label{app:vlm_eval}
To evaluate the semantic consistency of synthetic facial images, we used a multimodal large language model (VLM) to generate structured phenotype reports for real and DreamBooth-generated images. This analysis assesses whether synthetic faces reflect clinically meaningful traits of their target disease class. We excluded FastGAN images, as they are generated unconditionally and lack one-to-one correspondence with real patients. Since the VLM-based evaluation depends on matched image pairs for assessing regional semantic consistency, it is not applicable to class-agnostic generative models.

\begin{figure}[H]
    \centering
    \includegraphics[width=0.75\textwidth]{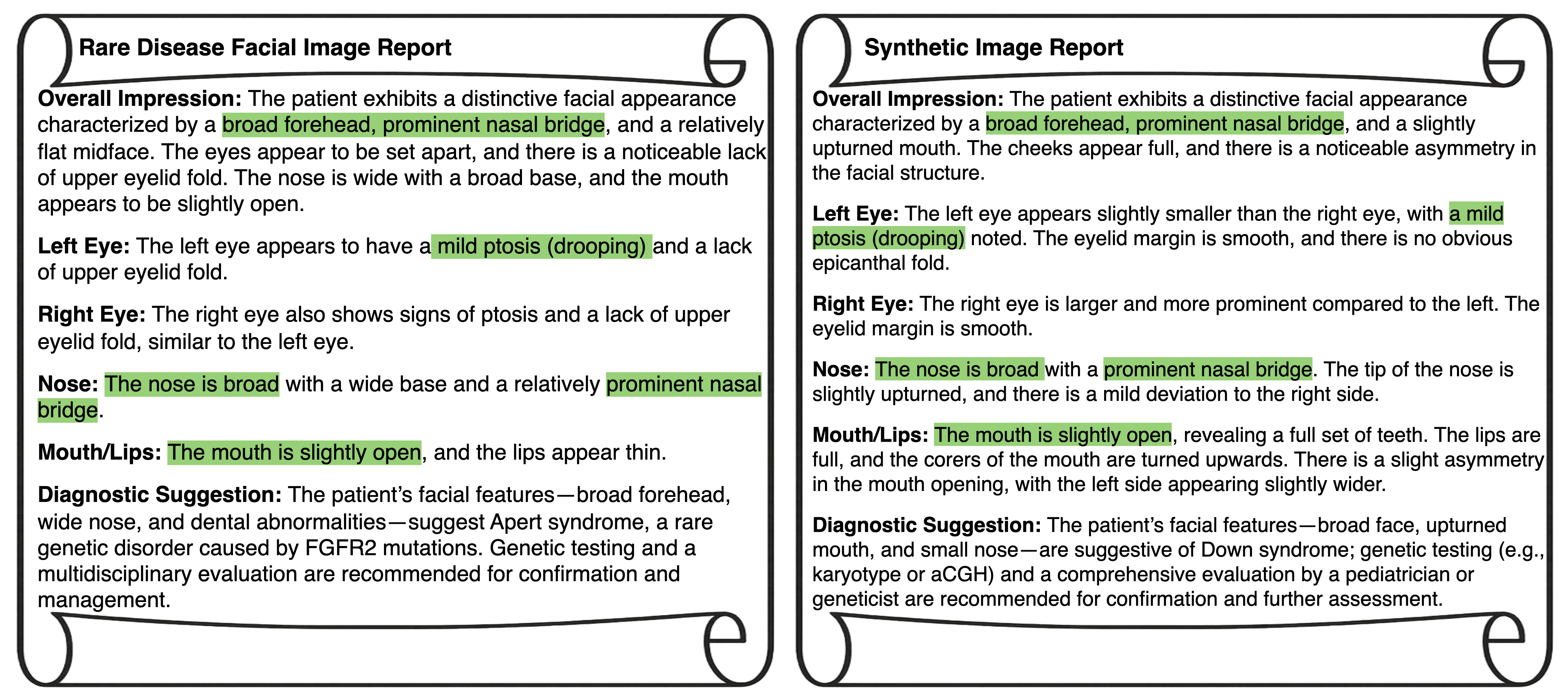}
    \caption{\textbf{High-Similarity Case.} The real and synthetic reports both describe ptosis, a broad nose, and slighly open mouth appearance. Minor deviations in phrasing are present, but the overall diagnostic impression remains aligned.}
    \label{fig:report_1}
\end{figure}

\begin{figure}[htbp]
    \centering
    \includegraphics[width=0.75\textwidth]{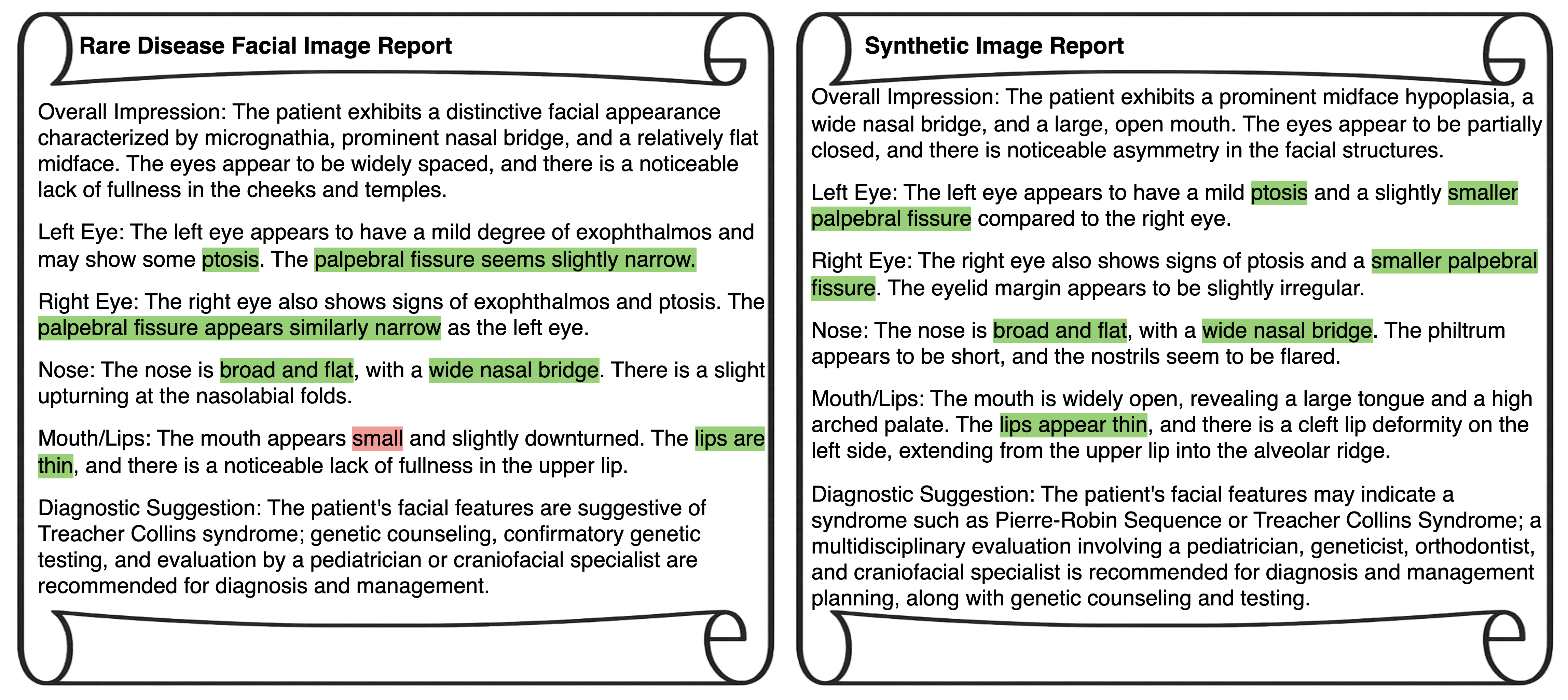}
    \caption{\textbf{High-Similarity Case.} Side-by-side comparison of phenotype reports from a real and synthetic image of the same disease class. The synthetic report closely mirrors the real one, especially in descriptions of palpebral fissures, ptosis, nasal shape, and lip structure.}
    \label{fig:report_3}
\end{figure}

\begin{figure}[htbp]
    \centering
    \includegraphics[width=0.75\textwidth]{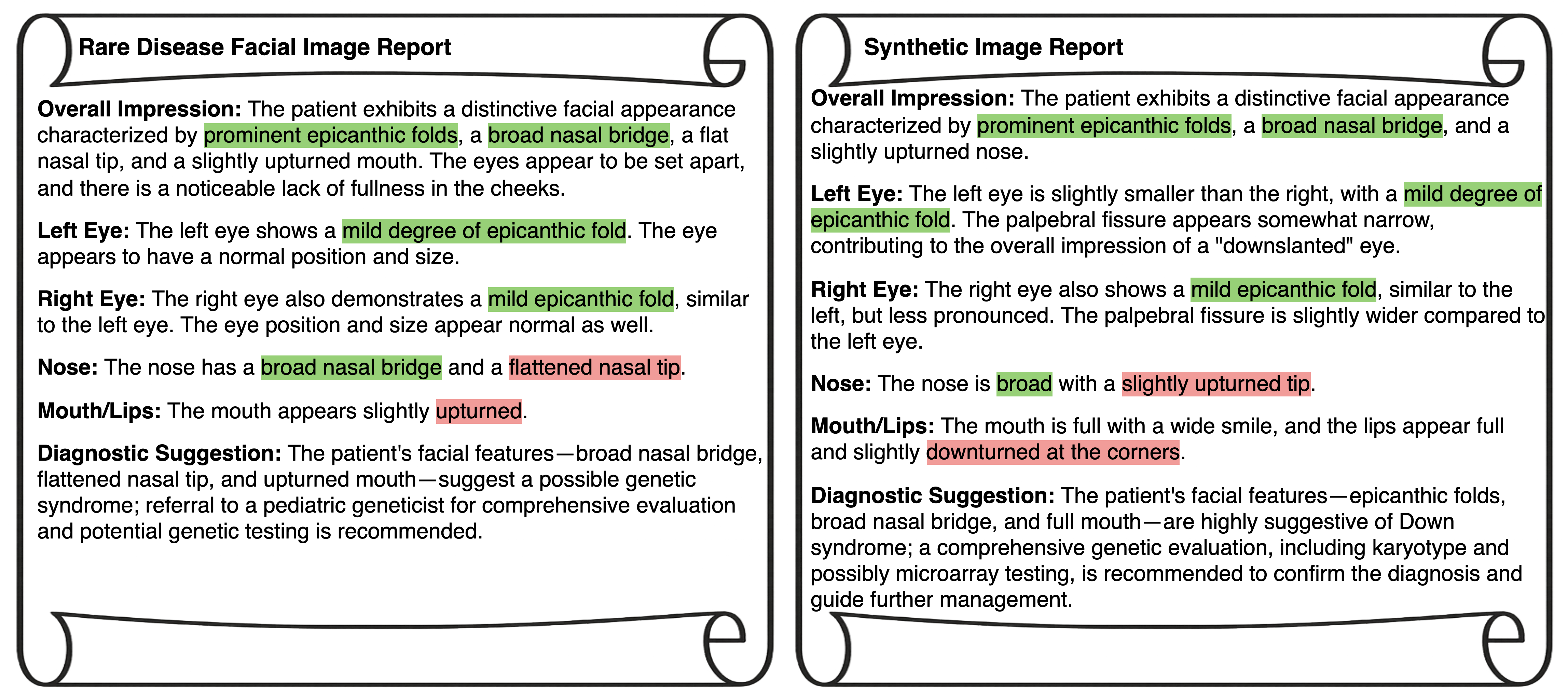}
    \caption{\textbf{Mixed-Agreement Case.} Although there is some alignment in features like nasal shape and bridge width, the synthetic report diverges in descriptors of lip configuration and diagnostic suggestion, showing partial inconsistency.}
    \label{fig:report_4}
\end{figure}

\begin{figure}[H]
    \centering
    \includegraphics[width=0.75\textwidth]{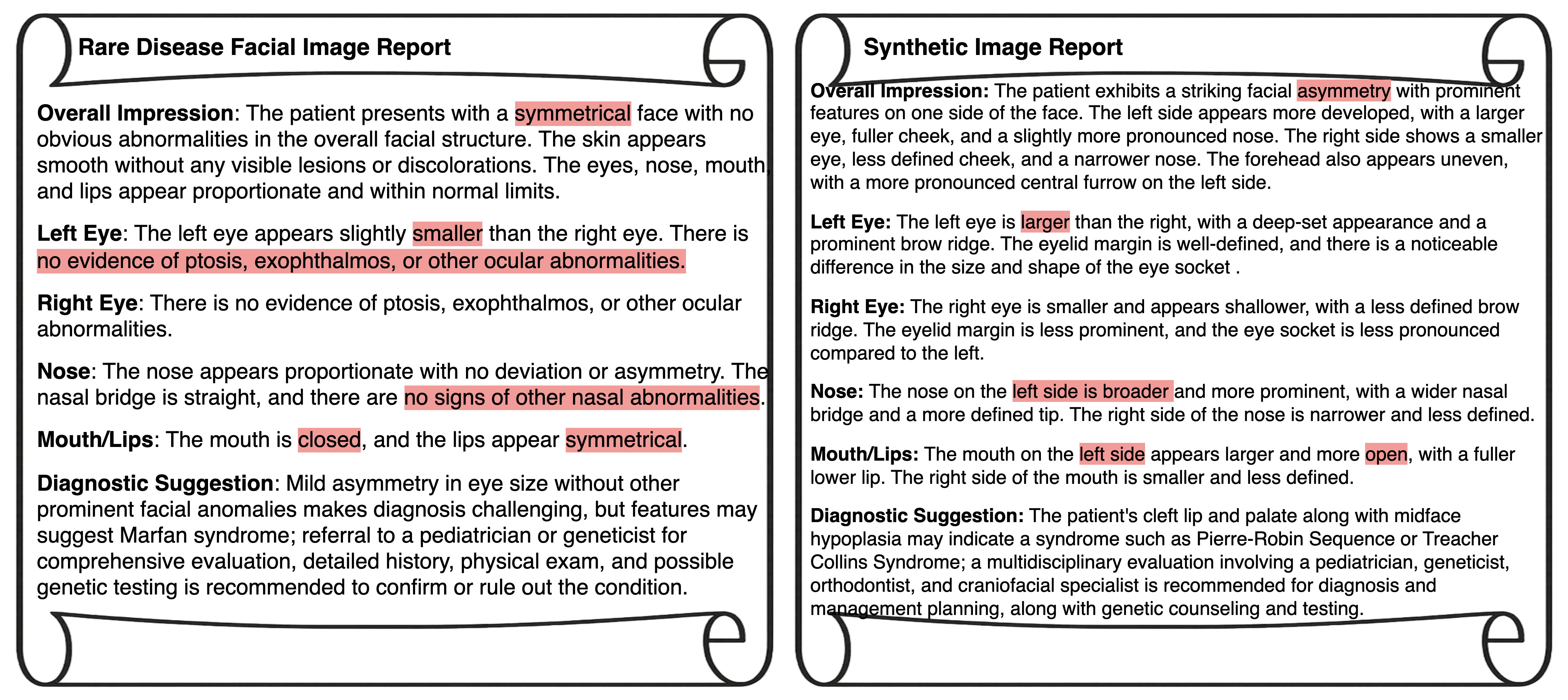}
    \caption{\textbf{Low-Similarity Case.} The real report describes normal symmetry, while the synthetic report notes asymmetry and deviation, showing the VLM interprets synthetic structure differently.}
    \label{fig:report_5}
\end{figure}

To illustrate this evaluation process, we present selected examples spanning high and low semantic similarity scores. For each example, we compare the phenotype reports generated from a real image and a DreamBooth-generated synthetic image of the same disease class. Region-level descriptions are extracted from both reports. Matching phenotype terms are highlighted in green, while contradictory or inconsistent terms are shown in red. The examples cover a range of semantic similarity levels, allowing qualitative inspection of both faithful and divergent generations. Detailed comparisons are shown in \Cref{fig:report_1}, ~\ref{fig:report_3}, ~\ref{fig:report_4}, and ~\ref{fig:report_5}.

\clearpage
\paragraph{BioBERT semantic similarity analysis}
We report the detailed results of the semantic similarity analysis between real and synthetic phenotype descriptions. Similarity scores were computed using BioBERT embeddings across five facial regions and an overall report segment. \Cref{tab:semantic_results} summarizes the mean and standard deviation of cosine similarities across the dataset. The highest alignment was observed in the nose and eye regions, while the mouth/lips showed slightly greater variability.

\paragraph{TF-IDF comparison}  
To complement the analysis presented above, we conducted a parallel evaluation using a traditional text similarity method based on TF-IDF \cite{manning2009} (\Cref{tab:semantic_results}). For each facial region and the overall report, we computed cosine similarity between real and synthetic descriptions using scikit-learn \cite{scikit-learn}'s \texttt{TfidfVectorizer}.

\renewcommand{\arraystretch}{1}
\setlength{\tabcolsep}{10pt}  
\begin{table}[htbp]
\centering
\caption{Semantic similarity and TF-IDF-based semantic similarity across five facial regions.}
\label{tab:semantic_results}
\begin{tabular}{lcc}
\toprule
\textbf{Region} & \textbf{BioBERT Similarity Score} & \textbf{TF-IDF Similarity Score} \\ 
\midrule
Overall     & 0.8404 (0.0748) & 0.7630 (0.0707) \\
Left Eye    & 0.7485 (0.1228) & 0.4364 (0.1358) \\
Right Eye   & 0.7535 (0.1423) & 0.4578 (0.1352) \\
Nose        & 0.7712 (0.1344) & 0.4315 (0.1432) \\
Mouth/Lips  & 0.7355 (0.1376) & 0.4612 (0.1321) \\
\bottomrule
\end{tabular}
\end{table}

These findings support the need for domain-specific semantic models when evaluating medical text. TF-IDF–based comparisons fail to fully capture conceptual similarity in phenotype language. While overall similarity trends are consistent, the lower region-wise scores highlight the limitations of surface-level lexical methods when applied to clinical narratives.

\subsection{Uncertainty and robustness analysis}
\label{app:vlm_cross_model}
\begin{figure}[H]
\centering
\includegraphics[width=0.5\textwidth]{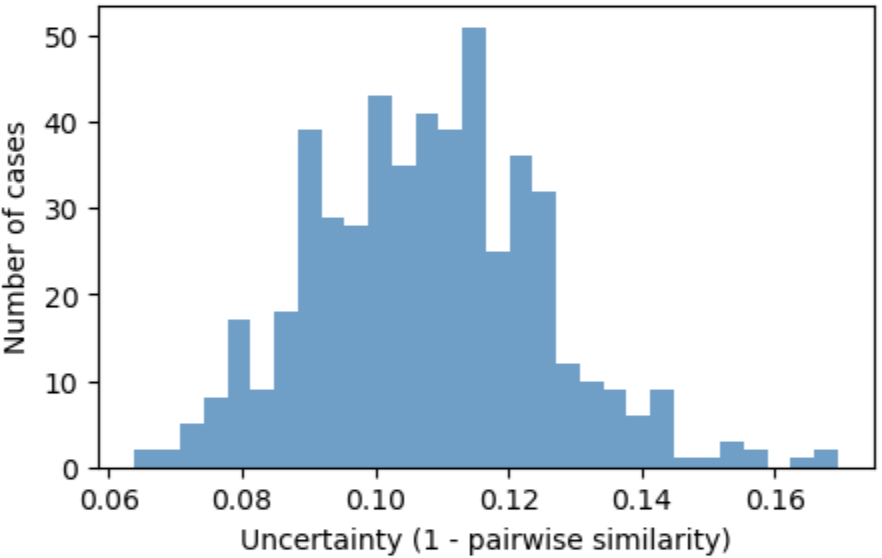}
\caption{\textbf{Distribution of uncertainty scores across all cases.}}
\label{fig:uncertainty_hist}
\end{figure}

\Cref{fig:uncertainty_hist} complements the uncertainty analysis in the main text by visualizing the distribution of uncertainty scores across all samples. 75\% of the cases exhibit low uncertainty ($<$ 0.14), indicating stable and consistent phenotype descriptions under stochastic sampling.

To further assess robustness, \Cref{tab:cross_model} reports cross-model phenotype similarity between Qwen2.5-VL and LLaVA-NeXT. The results show consistent similarity ranges across real and synthetic images, supporting that the evaluation is not sensitive to the choice of VLM.

\begin{table}[H]
\centering
\caption{Cross-model phenotype similarity.}
\label{tab:cross_model}
\begin{tabular}{lcc}
\toprule
 & \textbf{LLaVA (Real)} & \textbf{LLaVA (Synthetic)} \\
\midrule
\textbf{Qwen (Real)}      & 0.7053 (0.0711) & 0.7176 (0.0691) \\
\textbf{Qwen (Synthetic)} & 0.7204 (0.0726) & 0.7355 (0.0730) \\
\bottomrule
\end{tabular}
\end{table}

\clearpage
\section{Regional analysis and potential bias}
\label{app:regional_bias}

To assess potential bias in both classification performance and synthetic data evaluation, we analyze model behavior across geographic regions. However, population-level demographic attributes (e.g., ethnicity, ancestry, or skin tone) are not available in our dataset, as web-scraped rare disease case reports rarely provide standardized annotations. As a proxy, we stratify the data by geographic region (the only consistently recoverable attribute) and group samples into four regions: \textbf{Africa}, \textbf{Americas}, \textbf{Asia}, and \textbf{Europe}.

\begin{figure}[H]
    \centering
    \begin{subfigure}[t]{0.4\textwidth}
        \centering
        \includegraphics[width=\textwidth]{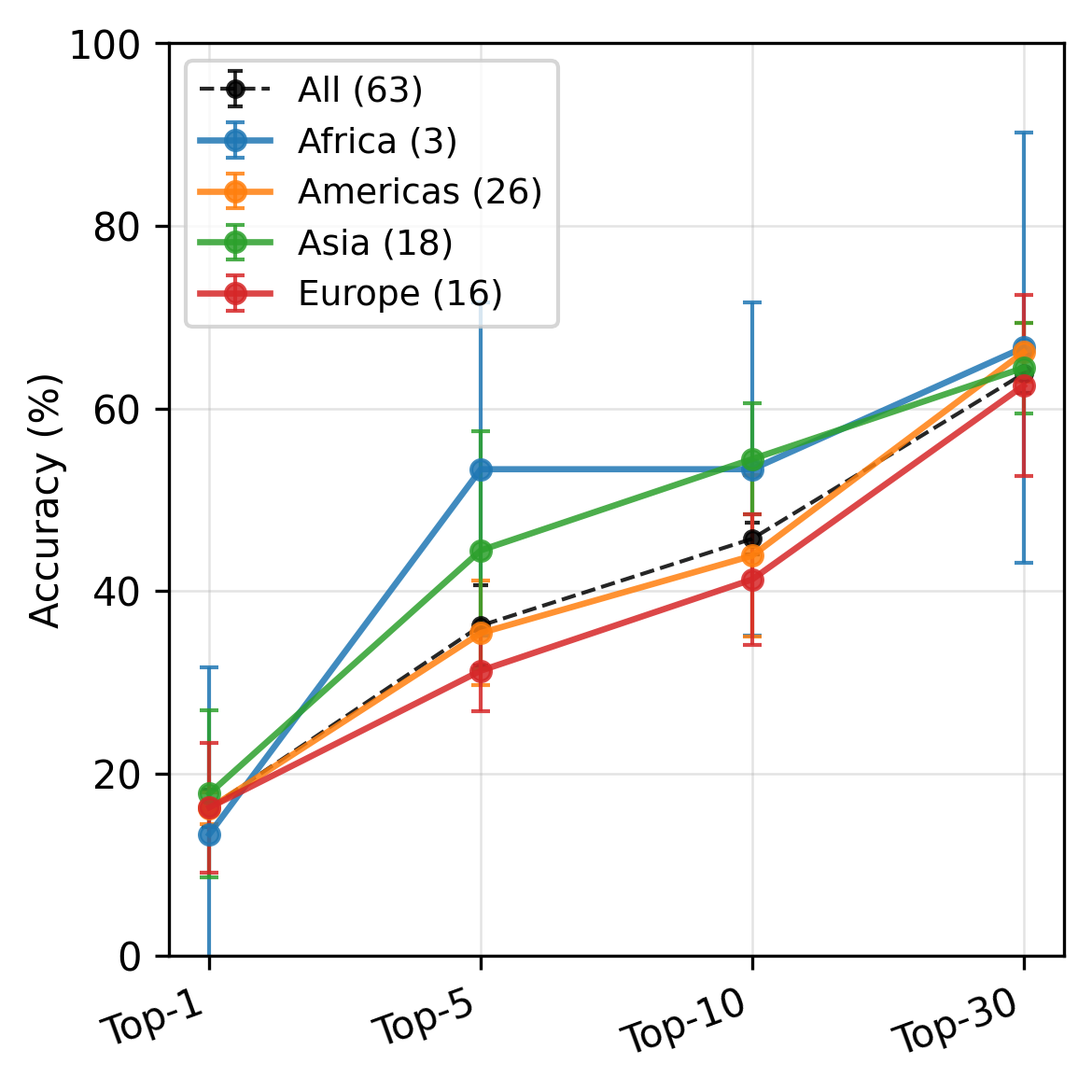}
        \caption{Regional diagnosis results across Top-$k$ accuracies.}
        \label{fig:region_topk}
    \end{subfigure}
    \hspace{0.06\textwidth}
    \begin{subfigure}[t]{0.4\textwidth}
        \centering
        \includegraphics[width=\textwidth]{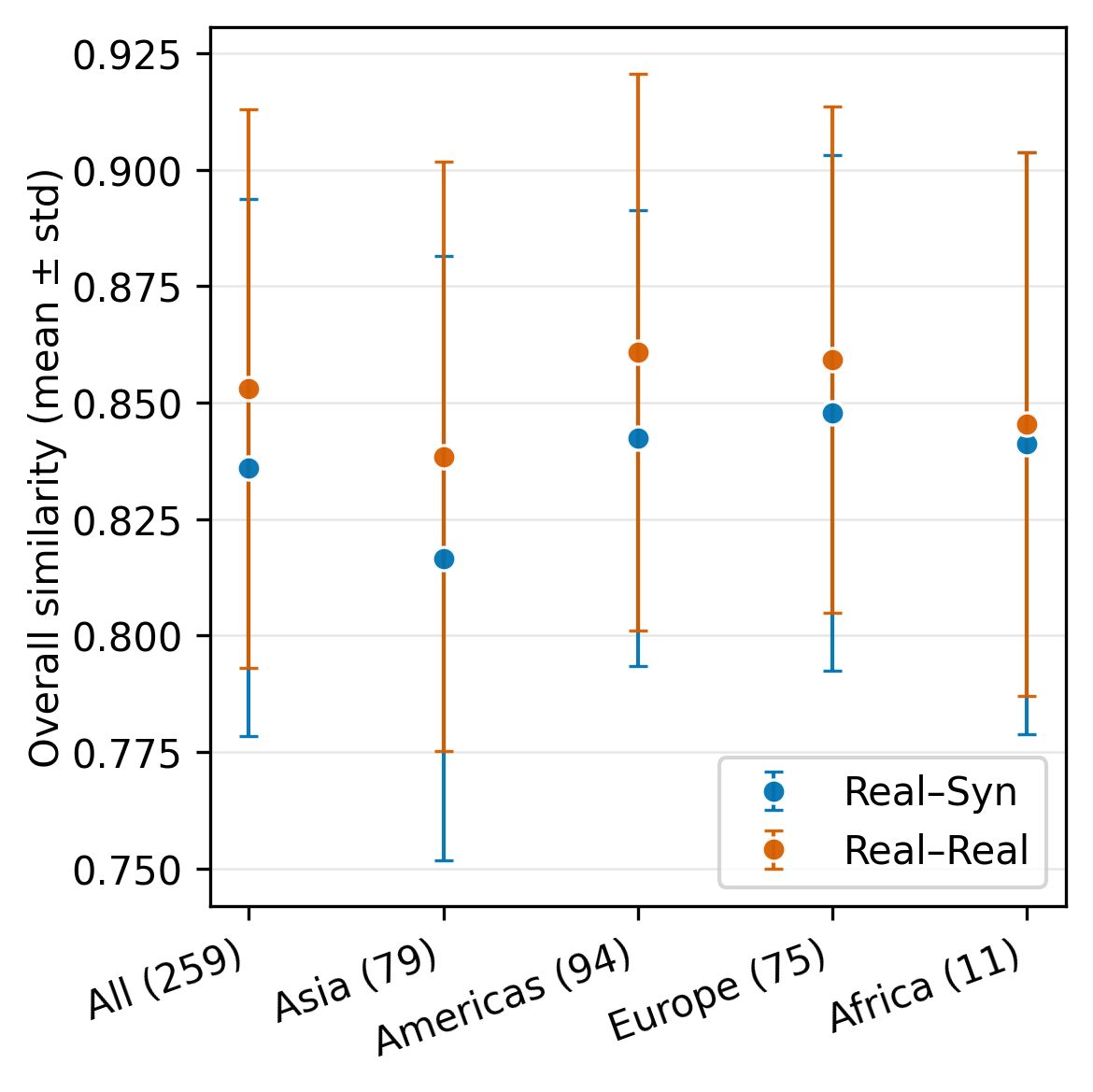}
        \caption{Regional phenotype similarity comparison.}
        \label{fig:region_similarity}
    \end{subfigure}

    \caption{\textbf{Regional analysis across geographic groups.}}
    \label{fig:regional_analysis}
\end{figure}

\paragraph{Regional diagnosis performance.}
We report Top-$k$ accuracy for supervised learning across regions in \Cref{fig:region_topk}. While minor variations are observed at lower $k$, performance trends are broadly consistent across regions and converge as $k$ increases, indicating similar generalization behavior as the results in main text. 

\paragraph{Regional phenotype similarity.}
We further evaluate phenotype similarity across regions using VLM-generated reports (\Cref{fig:region_similarity}). Real–synthetic similarity is comparable to real–real similarity across all regions within statistical uncertainty, suggesting that synthetic data preserves phenotype characteristics without introducing substantial regional bias. Notably, the Americas region shows slightly higher similarity scores, which may reflect a larger representation of cases from this region in the dataset. However, the overall consistency across regions supports the generalizability of our findings.

\paragraph{Limitations.}
We note that geographic region is only a coarse proxy and does not directly correspond to demographic attributes such as skin tone or ancestry. In addition, potential collection bias in publicly available case imagery may influence both classification and generative models. Future work should prioritize the collection of more diverse and well-annotated datasets to enable more granular analysis of demographic bias and ensure equitable performance across all patient populations and the development of synthetic data generation methods that explicitly account for demographic diversity.

\clearpage
\section{Statistical reporting details}
\label{app:stats_reporting}
All reported results in this paper include 1-sigma error bars, expressed as the standard deviation in parentheses (e.g., 6.90 (1.45)), computed over 5-fold cross-validation. The primary source of variability is the random train/test split across folds. For each configuration, we evaluate performance on the held-out fold and report the sample mean and standard deviation across the five runs. We assume approximately normally distributed metrics across folds, which is common in classification evaluation under low-data regimes. No additional sources of randomness (e.g., weight initialization or stochastic sampling) are varied unless explicitly stated.

Standard deviation is calculated using the unbiased estimator (i.e., Bessel’s correction):
\[
\mathrm{std}(x_1, \ldots, x_n) = \sqrt{\frac{1}{n-1} \sum_{i=1}^{n}(x_i - \bar{x})^2}
\]
This computation is implemented via NumPy \cite{harris2020}’s \texttt{np.std(..., ddof=1)} function.

\section{Hyperparameter settings and training details}
\label{app:training}

We summarize the key hyperparameters used in our benchmark experiments across three experimental settings in \Cref{tab:s2_classification},~\ref{tab:s3_fewshot},~\ref{tab:s4_dreambooth}, and~\ref{tab:s5_fastgan}. 

\subsection{Standard supervised classification}
\begin{table}[H]
\centering
\caption{Hyperparameters for standard supervised classification experiments.}
\label{tab:s2_classification}
\begin{tabular}{ll}
\toprule
\textbf{Parameter} & \textbf{Value} \\
\midrule
Optimizer & Adam \\
Learning rate & $1e^{-4}$ \\
Batch size & 32 \\
Training folds & 5-fold class split \\
Number of epochs & 50 \\
Loss function & CrossEntropyLoss \\
Weight decay & $1e^{-4}$ \\
Image size & $224 \times 224$ \\
Pretrained backbone & ImageNet for ResNet, DenseNet, Swin-T, CLIP  \\
 & VggFace2 for FaceNet \\
 & VggFace for VGG \\
\bottomrule
\end{tabular}
\end{table}

\subsection{Few-Shot learning}
\begin{table}[H]
\centering
\caption{Hyperparameters for Prototypical Networks.}
\label{tab:s3_fewshot}
\begin{tabular}{ll}
\toprule
\textbf{Parameter} & \textbf{Value} \\
\midrule
Optimizer & Adam \\
Learning rate & $1e^{-3}$ \\
Batch size & 1 \\
Training folds & 5-fold class split \\
Episodes per fold & 600 train / 100 val / 200 test \\
Loss function & CrossEntropyLoss \\
Distance metric & Euclidean Distance\\
\bottomrule
\end{tabular}
\end{table}

\subsection{Synthetic data generation}
\begin{table}[H]
\centering
\caption{Training settings for DreamBooth per disease class.}
\label{tab:s4_dreambooth}
\begin{tabular}{ll}
\toprule
\textbf{Parameter} & \textbf{Value / Description} \\
\midrule
Base model & SG161222/Realistic\_Vision\_V5.1\_noVAE \\
Training setting & Class-conditioned (per disease) \\
Text prompt & \texttt{"a child with [DISEASE] disease"} \\
Training steps & 800 per class \\
Batch size & 1 \\
Learning rate & $1e^{-6}$ \\
Image resolution & $512 \times 512$\\
Mixed precision & fp16 \\
Safety checker & NSFW filter \\
\bottomrule
\end{tabular}
\end{table}

\begin{table}[H]
\centering
\caption{Training settings for FastGAN on pooled real images.}
\label{tab:s5_fastgan}
\begin{tabular}{ll}
\toprule
\textbf{Parameter} & \textbf{Value / Description} \\
\midrule
Architecture & Original FastGAN (skip-layer excitation) \\
Training setting & Unconditional (no class labels) \\
Iterations & 80,000 \\
Batch size & 8 \\
Optimizer & Adam \\
Learning rate & $2e^{-4}$ \\
Image resolution & $512 \times 512$ \\
\bottomrule
\end{tabular}
\end{table}

\subsection{Hardware and compute resources}
Standard supervised classification, few-shot learning, DreamBooth, and FastGAN models were all implemented in PyTorch \cite{paszke2019} and trained using a single NVIDIA A100 GPU with 80GB of memory (also used for VLM-based report generation). Specifically, standard supervised classification and few-shot learning required approximately 240 and 550 minutes of training time, respectively, for downstream analysis. DreamBooth required 30–50 minutes of training per class, depending on the number of input images and prompt complexity, while FastGAN training took approximately 15 hours to converge.







